\documentclass{article}

\usepackage{arxiv}
\usepackage{fullpage}
\usepackage{charter}
\usepackage[round]{natbib}
\renewcommand{\bibname}{References}
\renewcommand{\bibsection}{\subsubsection*{\bibname}}

\usepackage{amsmath,amsfonts,bm}

\def\Figref#1{Figure~\ref{#1}}

\def\Secref#1{Section~\ref{#1}}

\def\eqref#1{equation~\ref{#1}}
\def\Eqref#1{Equation~\ref{#1}}

\def\1{\bm{1}}

\DeclareMathAlphabet{\mathsfit}{\encodingdefault}{\sfdefault}{m}{sl}
\SetMathAlphabet{\mathsfit}{bold}{\encodingdefault}{\sfdefault}{bx}{n}

\DeclareMathOperator*{\argmax}{arg\,max}
\DeclareMathOperator*{\argmin}{arg\,min}

\newcommand{\framework}{\textsc{Model Selector}}

\usepackage[T1]{fontenc}
\usepackage[utf8]{inputenc}
\usepackage{babel}
\usepackage{hyperref}
\usepackage{url}
\usepackage{longtable}
\usepackage{cleveref}
\usepackage{nicefrac}
\usepackage{booktabs}
\usepackage{mathtools}
\usepackage{multirow}
\usepackage{algorithm}    
\usepackage{algpseudocode} 
\usepackage{bbm}
\usepackage{amsfonts}
\usepackage{dsfont}

\usepackage{titlesec}

\titlespacing{\section}{0pt}{*0.5}{*0.5} 
\titlespacing{\subsection}{0pt}{*0.5}{*0.5} 

\usepackage{arxiv}
\usepackage{fullpage}
\usepackage{charter}

\usepackage[utf8]{inputenc} 
\usepackage[T1]{fontenc}    
\usepackage{hyperref}       
\usepackage{url}            
\usepackage{booktabs}       
\usepackage{amsfonts}       
\usepackage{nicefrac}       
\usepackage{microtype}      
\usepackage{lipsum}
\usepackage{fancyhdr}       
\usepackage{graphicx}       
\graphicspath{{media/}}     

\usepackage{titlesec}
\titlespacing*{\section}{0pt}{12pt}{6pt}
\titlespacing*{\subsection}{0pt}{8pt}{4pt}

\usepackage{amsmath}
\usepackage{hyperref}
\hypersetup{hidelinks}
\usepackage{graphicx}
\usepackage{enumitem}
\usepackage{setspace}
\usepackage{wrapfig}
\usepackage{multirow}
\usepackage{float}
\usepackage{stfloats}
\usepackage{xcolor}
\title{All models are wrong, some are useful:\\ Model Selection with Limited Labels \vspace{1em}}

\author{
Patrik Okanovic \\
  ETH Zurich \\
  \texttt{patrik.okanovic@inf.ethz.ch} \\ \\ \\
\textbf{Torsten Hoefler} \\
ETH Zurich \\
\texttt{htor@ethz.ch} \\ \\
  \And
Andreas Kirsch  \\
  \texttt{blackhc@gmail.com} \\ \\ \\ \\
\textbf{Andreas Krause} \\
  ETH Zurich \\
  \texttt{krausea@ethz.ch} \\
  \And
Jannes Kasper \\
  TU Delft \\
  \texttt{J.Kasper@student.tudelft.nl} \\ \\ \\
\textbf{Nezihe Merve Gürel} \\
  TU Delft \\
  \texttt{n.m.gurel@tudelft.nl} \\
}
\begin{document}
\maketitle

\begin{abstract}
\noindent 
We introduce \framework{}\footnote{Source code: \url{https://github.com/RobustML-Lab/model-selector}}, a framework for label-efficient selection of pretrained classifiers. 
Given a pool of unlabeled target data, \framework{} samples a small subset of highly informative examples for labeling, in order to efficiently identify the \emph{best} pretrained model for deployment on this target dataset. 
Through extensive experiments, we demonstrate that \framework{} drastically reduces the need for labeled data while consistently picking the best or near-best performing model. 
Across 18 model collections on 16 different datasets, comprising over 1,500 pretrained models, \framework{} reduces the labeling cost by up to $94.15\%$ to identify the best model compared to the cost of the strongest baseline. 
Our results further highlight the robustness of \framework{} in model selection, as it reduces the labeling cost by up to $72.41\%$ when selecting a near-best model, whose accuracy is only within $1\%$ of the best model.

\end{abstract}
\section{Introduction}
\looseness -1 The abundance of openly available machine learning models poses a dilemma: selecting the best pretrained model for domain-specific applications becomes increasingly challenging.
As the development and deployment of large-scale machine learning models have been accelerating at a rapid pace~\citep{he2016deep,chiu2018state,mann2020language,radford2019language},
a wide selection of pretrained models for natural language processing~\citep{devlin2019bertpretrainingdeepbidirectional,liu2019robertarobustlyoptimizedbert} and computer vision~\citep{NIPS2012_c399862d,xie2017aggregatedresidualtransformationsdeep}, varying in architecture, type, and complexity, are now accessible through various open-source and academic platforms~\citep{huggingface,pytorchhub,tensorflowhub}.
AutoML platforms~\citep{aws,googlecloud} 
further increase this variety by providing researchers and developers with instant access to powerful bespoke models through automated workflows and pretrained model repositories.
Thanks to advances in training methods and model architecture, many of these models now support zero-shot learning~\citep{brown2020language,xian2018zero,pourpanah2022review,radford2021learningtransferablevisualmodels}, and can tackle new tasks without requiring fine-tuning or updates to their weights.
Traditionally, models are often picked by hand, involving intuition-based sampling and evaluation, which quickly becomes impractical as the number of models and potential evaluation samples increases, risking suboptimal choices and inefficient resource utilization.

To address this, several techniques perform automated model validation by selecting and labeling a small informative subset of data examples to assess model performance and facilitate label-efficient model selection and evaluation~\citep{NIPS2012_92fb0c6d,matsuura2023active,kossen2021active}, which are also commonly referred to as \emph{active model selection}~\citep{karimi2021online,matsuura2023active,9101367,gardner2015bayesian}. 
However, many of the methods in the pool-based setting (characterized by the availability of a large pool of unlabeled data) either assume a fixed number of available models~\citep{NIPS2012_92fb0c6d} or are restricted to specific model families~\citep{gardner2015bayesian}. Others require retraining the model each time a new example is labeled~\citep{ali2014active}, which can be computationally expensive and potentially unnecessary for ready-to-deploy models. 

In this work, we aim to extend these settings and ask: \emph{Given a pool of unlabeled data, how can we identify the most informative examples to label in order to select the best classifier for this data, \emph{both in a model-agnostic and efficient manner}?}
\begin{figure*}[t]
    \centering
    \setlength{\abovecaptionskip}{2pt}
    \includegraphics[width=\linewidth]{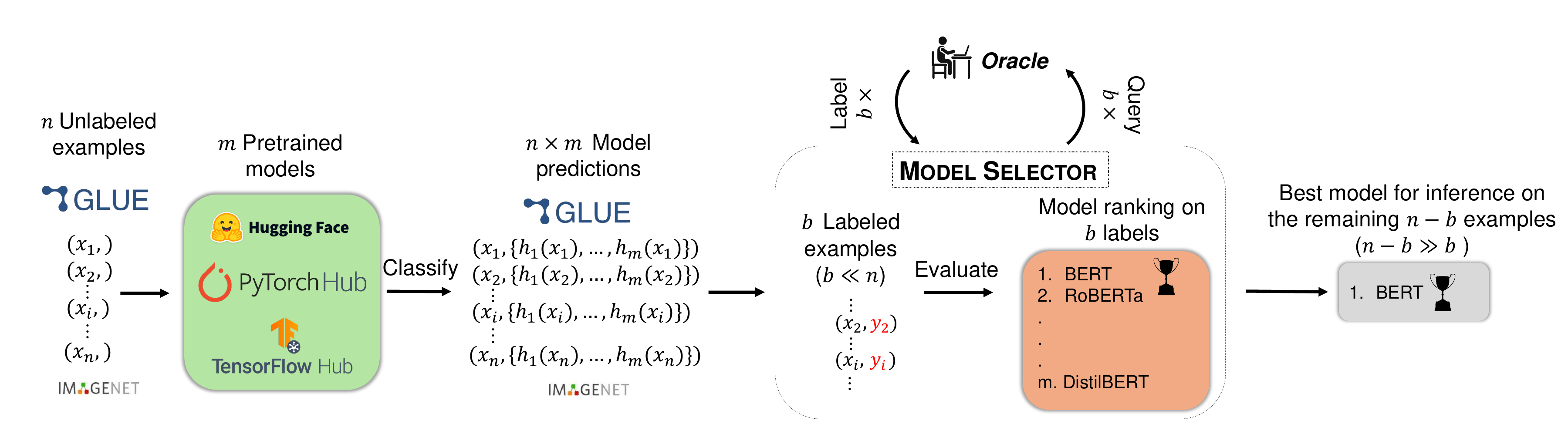}
    \caption{An overview of our label-efficient model selection pipeline with \framework{}. Given a pool of $n$ unlabeled data examples and a set of $m$ pretrained classifiers, \framework{} aims to select $b$ (with $b\ll n$) unlabeled examples that, once labeled, can identify the best pretrained model.}
    \label{fig:overview}
\end{figure*}
\paragraph{Contributions} In this paper, we introduce \framework, a fully \emph{model-agnostic} approach for identifying the best pretrained model using a limited number of labeled examples. Unlike existing methods, which often rely on assumptions about the model architectures or require detailed knowledge of model internals, our approach treats models purely as black-boxes requiring no additional training and anything beyond hard predictions, i.e.,~the predicted label without predictive distribution. 
To the best of our knowledge, this is the first work in the pool-based setting that is both model-agnostic and relies solely on hard predictions.

Formally, given a pool of $n$ newly collected, unlabeled data examples and a labeling budget $b$ with $b\ll n$, \framework\ aims to select $b$ unlabeled examples that, once labeled, provide the maximum expected information gain about the best model (model with the highest utility) for this data. 
To achieve this, we employ the \emph{most informative selection policy}~\citep{chen2015sequential,kirsch2019batchbaldefficientdiversebatch,treven2023optimistic,chattopadhyayperformance,ding2024learning}, where we greedily label the most informative examples until the labeling budget $b$ is exhausted. We define informativeness in terms of mutual information between the best model and data labels, relying on a \emph{single-parameter} model as proxy that nonetheless sufficiently captures the relationship between them. 
Despite its simplicity, the proposed single-parameter model is highly effective in identifying the best model under limited labels. Moreover, for such a simplified model, the most informative selection policy has a provable near-maximal utility, as shown by~\citet{chen2015sequential}.

We conduct extensive experiments comparing our method against a range of adapted methods over $1,500$ models across $18$ different model collections and $16$ different datasets. We observe that \framework{} consistently outperforms all the baselines. 
In particular, \framework{} reduces the labeling cost by up to $94.15\%$ compared to the best competing baseline in finding the best model.
We also show that \framework{} finds a near-best model with a reduction in labeling cost by up to $72.41\%$ within $1\%$ accuracy of the best model.

Once the best pretrained machine learning model is identified, we then deploy it for making predictions on the remaining unlabeled samples. An overview of \framework{} is provided in \Figref{fig:overview}.
Our approach is designed to reduce the expenses related to pretrained model selection for classifiers, marking a step forward in efficient machine learning practice.

\section{Related Work}
\label{sec:related_work}
To date, label-efficient model selection has been studied primarily in the stream-based (online) setting~\citep{piratla2021active,liu2022contextual,kassraie2023anytime,madani2012activemodelselection,xia2024convergenceaware,liu2022cost,li2024necessity,li2024online,xia2024llm,karimi2021online}. In the pool-based setting, where we have access to the entire collection of unlabeled examples, the literature often makes strong assumptions, such as predefined model families or learning tasks. For example, both ~\citet{9101367} and ~\citet{zhao2008active} assume specific learning tasks (binary time series classification or graph-based semi-supervised learning, respectively), while ~\citet{gardner2015bayesian} limit model selection to Gaussian processes. Other works including~\citet{ali2014active} consider a general class of hypotheses but rely on sequential training. \citet{bhargav2024submodular} introduce a misclassification penalty framework for hypothesis testing in a batched setting, and~\citet{kumar2018classifier} perform classifier risk estimation on stratified batches without strong assumptions.

Several works align closely with our setting, focusing on model selection without the previously mentioned assumptions. These works focus on evaluating the risk of a single model~\citep{sawade2010active,6413890,kossen2021active}, comparing two models~\citep{NIPS2012_92fb0c6d,leite2010active}, or, like ours, comparing multiple models~\citep{matsuura2023active}. We adapt the method of~\cite{NIPS2012_92fb0c6d} for pairwise model comparison in our setting (\textsc{amc}) and use \textsc{vma}~\citep{matsuura2023active} as a baseline. While \textsc{amc} minimizes the asymptotic variance of the estimated loss for each model, \textsc{vma} minimizes the variance of the estimated test loss conditioned on previously queried examples.
Due to the lack of applicable baselines, we also employ uncertainty and margin sampling~\citep{dagan1995committee,seung1992query,freund1997selective}. 

Unlike previous works, we propose a fully agnostic method, making no assumptions about model families, tasks, or the use of soft predictions. Additionally, our work is related to optimal information-gathering strategies, with a discussion of related work deferred to the \Cref{sec:app_rel_work}, in order to prioritize works with the same objective of label-efficient model selection.

\section{Model Selector}
\label{sec:method}
In this section, we present our label-efficient model selection algorithm, \framework{}. We start by describing the problem setup, where we connect the unknown best model to the true data labels using a single-parameter likelihood model. Then, we introduce the \framework{} algorithm, which is designed to identify the best pretrained model with limited labels. We end the section by providing further details on the algorithm and explain how this single parameter can be learned directly from the data without requiring any labels.

\subsection{Problem Setting}  
Consider a pool of $n$ freshly collected unlabeled examples denoted by $\mathcal{D}=\{(x_i, Y_i) \in \mathcal{X}\times \mathcal{Y}\ | {i\in[n]}\}$, whose labels $Y_i$ are unobserved\footnote{In here and what follows $[n] \coloneq \{1, 2, ...n\}$.}. We denote the true labels by $y_i$.


For a given set of $m$ pretrained, ready-to-deploy classifiers $\mathcal{M}=\{h_j: \mathcal{X}\mapsto \mathcal{Y}\ |j\in[m]\}$, and a labeling budget $b$ with $b\ll n$, our aim is to identify the best pretrained model $h(\cdot)$ among $\mathcal{M}$ by querying a small number of at most $b$ labels, and perform inference for the remaining $n-b$ examples using this model. 
The \emph{best} model here is defined as the model that fits the data $\mathcal{D}$ best if all $n$ labels $y_{i\in[n]}$ were available. In this work, we consider the model with the highest utility (accuracy) on $n$ labels as the best. 
That is, $h^* = \argmax_{h_j\in \mathcal{M}} \nicefrac{1}{n} \sum_{i=[n]} \mathds{1}[h_j(x_i)=y_i]$.

\looseness -1 The best model is unknown to us without labels.
We represent this unknown by using a random variable $H$, with a known prior distribution $H\sim \mathbb{P}(H=h_j)$ over the set of candidate models $\mathcal{M}=\{h_j\hspace{.2em}| j\in[m]\}$.
Our goal is to uncover the identity of $H$ for the target data $\mathcal{D}$ by using as few labels as possible. 
That is, we want $h^* = \argmax_{h_j\in\mathcal{M}}\mathbb{P}(H=h_j|\mathcal{L})$ where $\mathcal{L}$ denotes the observed labels.
To achieve this, we will sequentially label new data examples that reveals information about $H$.

Given that accuracy is our chosen utility measure, we focus on the binary outcome of correct versus incorrect predictions made by $H$ among the candidate models $\mathcal{M}$ on these labeled examples.
With this intention, we use a parameter that represents the probability of this binary outcome.
Formally, we characterize the behavior of the best model $h^*$ by an error probability $\epsilon$ when predicting the true label as follows:
\begin{align}\label{eqn:uncertainty-model}
    &\mathbb{P}(H(x)\neq y|H=h^*) = \epsilon,\\
    &\hspace{.2em} \mathbb{P}(H(x)= y|H=h^*) = 1-\epsilon\notag
\end{align}
where $\epsilon \in [0, 1]$. 
We also assume that the true (unobserved) labels $\{Y_i\hspace{.2em}|i\in[n]\}$ are conditionally independent given $H$. 
Thus, they can be interpreted as generated according to the best model $h^*$, and flipped with probability $\epsilon$ independently at random. We learn $\epsilon$ prior to model selection process, as detailed in~\Secref{sec:epsilon}.  

Our primary motivation for this formulation is to establish a highly compact yet effective and interpretable relationship between the best model and the true labels.
Although this problem setting closely aligns with the one studied by~\cite{chen2015sequential}, which inspired our approach, representing all class conditional relationships with a single parameter is a major simplification of the problem.
Next, we introduce the \framework{} algorithm and explain how it uses \Eqref{eqn:uncertainty-model} to select informative examples to label, followed by algorithmic details concerning the choice of $\epsilon$.

\subsection{The Algorithm} 
Given an unlabeled data pool $\mathcal{D}$, we want to identify the $b$ most informative examples to label for model selection. We characterize the information contained in the labels towards $H$ using Shannon's mutual information with respect to \Eqref{eqn:uncertainty-model}. 

Entropy, as a measure of uncertainty, given a discrete random variable $X$ distributed according to $\mathbb{P}: \mathcal{X} \rightarrow [0,1]$ is defined as $\mathbb{H}(X) := - \sum _{x \in \mathcal{X}} \mathbb{P}(X=x) \log \mathbb{P}(X=x)$. The mutual information between $X$ and $Y$ measures the expected information gain that $Y$ provides about $X$, and is defined as: $\mathbb{I} (X; Y) = \mathbb{H}(X) - \mathbb{H} (X | Y)$.  

Our objective can be formalized as finding the set of labeled data examples $\mathcal{L}_{\textrm{OPT}[b]}$ (of size at most $b$) that gives us maximum information about $H$. That is,
\begin{equation}\label{eqn:objective}
    \mathcal{L}_{\textrm{OPT}[b]}:= \argmax_{\substack{\mathcal{L}\subset \{(x_i, y_i)\hspace{0.1em} |i\in[n]\}\\s.t.\ |\mathcal{L}|\leq b}} \mathbb{I}(H; \mathcal{L}).
\end{equation}
Starting from the initial pool of unlabeled examples $\mathcal{D}$, the strategy of \framework\ is to greedily pick the unlabeled example that provides the maximal information gain about the true value of $H$ and request its label. At each greedy step $t$, \framework\ queries the label of $x_t$ where:
\begin{equation}\label{eqn:informative-sampling}
    x_t = \argmax_{x\in \mathcal{U}_{t}}\  \mathbb{I}(H; Y|x, \mathcal{L}_{t})
\end{equation}
where $\mathcal{L}_t$ and $\mathcal{U}_t$ respectively denote the disjoint sets of labeled and unlabeled examples at step $t$.

To compute the information gain for the unlabeled data example $x$, we can express \Eqref{eqn:informative-sampling} in terms of differential entropies as follows:
\begin{align}\label{eqn:entropy-sampling}
    x_t &= \argmax_{x\in \mathcal{U}_{t}} \ \mathbb{H}(H|\mathcal{L}_{t}) - \mathbb{E}_{Y} [\mathbb{H}(H| \mathcal{L}_{t}\cup \{(x, Y)\})]\notag\\
    &= \argmin_{x\in \mathcal{U}_{t}} \ \mathbb{E}_{Y} [\mathbb{H}(H| \mathcal{L}_{t}\cup \{(x, Y)\})]
\end{align}
where the expectation in $\mathbb{E}_{Y}[\cdot]$ is taken over $Y$, since the true label $y$ for the corresponding $x$ is unknown\footnote{To compute this expectation, we approximate the posterior over the classes as uniform over the model predictions, motivated by its robust performance observed in the results.}. 

\Eqref{eqn:entropy-sampling} is equivalent to minimizing the model posterior uncertainty~\citep{nguyen2021information}. It suggests that the most informative sampling policy for unlabeled data is equivalent to minimizing the entropy of the posterior of $H$. 
To compute this entropy, we require the expected entropy of the model posterior given the example $x$ and its hypothetical label $c\in\mathcal{Y}$ as well as the labeled examples $\mathcal{L}_t$, which we refer to as the \emph{hypothetical model posterior} at $t$ given $x$. By applying Bayes' rule, we obtain the following expression of the hypothetical model posterior:
\begin{align}\label{eqn:hypothetical-posterior}
    &\mathbb{P}(H=h_j|\mathcal{L}_{t}\cup \{(x, Y=c)\})\propto \\
    &\quad \mathbb{P}(\mathcal{L}_{t}\cup \{(x, Y=c)\})|H=h_j) \ \mathbb{P}(H=h_j).\notag
\end{align}
Consider a uniform prior over models with $P(H=h_j)=\nicefrac{1}{m}$ for simplicity. We have $\mathbb{P}(H=h_j|\mathcal{L}_{t}\cup \{(x, Y=c)\})\propto \mathbb{P}(\mathcal{L}_{t}\cup \{(x, Y=c)\})|H=h_j)$. Further applying the model in~\Eqref{eqn:uncertainty-model} to $\mathbb{P}(\mathcal{L}_{t}\cup \{(x, Y=c)\})|H=h_j)$, the hypothetical model posterior can be computed as:
\begin{align}\label{eqn:likelihood-given-model}
   &\mathbb{P}(H=h_j|\mathcal{L}_{t}\cup \{(x, Y=c)\}) \propto \\
   &\quad  (1-\epsilon)^{h_{j, (t, x)}}\epsilon^{t-h_{j, (t, x)}} \notag
\end{align}
where $h_{j, (t, x)}$ denotes the number of correct predictions of classifier $h_j$ on $\mathcal{L}_{t}\cup \{(x, Y=c)\})$.

Upon selecting the most informative example $x_t$ at round $t$ using \Eqref{eqn:entropy-sampling}, and receiving its label $y_t$ from the $oracle$, \framework\ updates the labeled set $\mathcal{L}_t$ with $\mathcal{L}_{t+1} = \mathcal{L}_t \cup \{x_t, y_t\}$ and unlabeled set $\mathcal{U}_t$ with $\mathcal{U}_{t+1} = \mathcal{U}_t \backslash \{x_t\}$ as well as the model posterior with 
\begin{align}\label{eqn:posterior-update}
    &\mathbb{P}(H=h_j|\mathcal{L}_{t+1}) \propto \\
    &\quad \mathbb{P}(H=h_j|\mathcal{L}_{t})(1-\epsilon)^{\mathds{1} [h_j(x_t)=y_t]}\epsilon^{\mathds{1}[h_j(x_t)\neq y_t]} \notag
\end{align}

It then selects another example to label from the pool of remaining unlabeled examples $\mathcal{U}_t$. After a number of these labels are requested up to the budget $b$, \framework\ returns the model with the highest accuracy on the labeled set $\mathcal{L}_b$. The pseudocode of this algorithm is depicted in \Cref{alg:general_alg}.
\begin{algorithm}[H]
    \caption{\framework{}}
    \label{alg:general_alg}
    \begin{algorithmic}[1]
    \small
    \Require models $\mathcal{M} = \{ h_1,\dots, h_m \}$, unlabeled examples $\mathcal{U}_0$ , parameter $\epsilon$, labeling budget $b$, $oracle$
    \State $\mathcal{L}_0 \gets \{ \}$
    \For{$t=0$ to $b-1$}
    \For{$c \in \mathcal{Y}$}
    \State $\mathbb{P}(H = h_j \mid \mathcal{L}_t \cup \{(x, Y=c)\}) \gets \frac{1}{Z}\mathbb{P}(H=h_j|\mathcal{L}_{t})
    (1-\epsilon)^{\mathds{1} [h_j(x_t)=c]}\epsilon^{\mathds{1}[h_j(x_t)\neq c]}$ \Comment{hypothetical model}
    \Statex  \hspace{41.5em}  \text{posterior} 
    \EndFor
    \State $x_t \gets \argmin_{x \in \mathcal{U}_t} \mathbb{E}_Y [ \mathbb{H} (H|\mathcal{L}_{t}\cup \{(x, Y)\}) ]$
    \State $y_t \gets oracle(x_t)$
    \State $\mathcal{L}_{t+1} \gets \mathcal{L}_t \cup \{ (x_t, y_t) \}$
    \State $\mathcal{U}_{t+1} \gets \mathcal{U}_t \backslash \{ x_t \}$
    \State $\mathbb{P}(H=h_j|\mathcal{L}_{t+1}) \gets \frac{1}{Z} \mathbb{P}(H=h_j|\mathcal{L}_{t}) (1-\epsilon)^{\mathds{1}[h_j(x_t)=y_t]} \epsilon^{\mathds{1}[h_j(x_t)\neq y_t]}$ \Comment{update model posterior}
    \EndFor \\
    \Return $\argmax_{h_j \in \mathcal{M}} \frac{1}{b} \sum_{i \in \mathcal{L}_b} \mathds{1}[h_j(x_i)=y_i]$ \Comment{select the best model}
\end{algorithmic}
\end{algorithm}

\subsection{Algorithmic Details}\label{sec:epsilon}
We determine the value of \(\epsilon\) prior to the model selection process by conducting a grid search over a range of values and choosing $\epsilon$ that yields the best performance in terms of accurately identifying the best model (with details deferred to \Secref{sec:eval-protocol}) \emph{without requiring labels}. 
In many active learning scenarios, it is a standard practice to allocate an initial budget for exploration~\citep{lewis1995sequential,mccallum1998employing,zhang2002active,hoi2006large,zhan2022comparative}, which typically involves randomly sampling some examples and querying their labels to obtain a rough estimate of dynamics of interest. This would yield an \(\epsilon\) for our work. 
However, upon exploring this option, we discovered that \(\epsilon\) can also be effectively learned by generating noisy labels using pretrained models. Specifically, we construct a noisy oracle, denoted as \(\{\hat{y}_i\}_{i \in [n]}\), for the unlabeled examples \(\{x_i\}_{i \in [n]}\) by leveraging the distribution of predicted classes from the candidate models. For tasks with a limited number of classes, we label each example with the most frequently predicted class; for tasks with a larger set of classes, we sample a class based on this distribution. We then carry out our grid search using these noisy labels as if they were the true oracle, and choose the best-performing $\epsilon$.

For all datasets and model collections, we estimate $\epsilon$ for our problem without the need for any initial labeled data. Quantitatively, our estimation of $\epsilon$ has an error margin of only $0.01$ compared to the values obtained using the ground truth oracle. We refer to \Cref{sec:app_epsilon} for more details on this process.

According to \Eqref{eqn:uncertainty-model}, an immediate expectation is that the ideal value for \(\epsilon\) should reflect the the error rate of the best model for the target data, guiding us to the data examples where our epistemic uncertainty about the best model is highest.
However, our experiments indicate that this can sometimes lead to overfitting to the evidence, possibly due to our single-parameter model in \Eqref{eqn:uncertainty-model} being a further simplification of the model proposed by~\cite{chen2015sequential} and misaligned with their modeling assumptions.
To illustrate this, consider a scenario in which the best model has an accuracy of $0.9$ on the target dataset, which implies that $\epsilon=0.1$. In this case, at each step, the probability that a model $h_j\in\mathcal{M}$ is the best model, represented by the posterior $\mathbb{P}(H=h_j |\mathcal{L}_t)$, is scaled by a factor of \( \nicefrac{(1-\epsilon)}{\epsilon} \) if $h_j$ correctly predicts the label for $x_t$, as specified in \Eqref{eqn:posterior-update}. For $\epsilon=0.1$, this scaling factor becomes $9$, which heavily weights the evidence and leads to overfitting.
Our grid search results support this observation: larger values of $\epsilon$ lead to more conservative updates to the model posterior, providing a regularization effect that mitigates overfitting to the evidence and improves overall performance.

The parameter $\epsilon$ can also be interpreted in terms of the model disagreement within the exploration-exploitation tradeoff. 
When $\epsilon$ is high, it yields better results in scenarios where the models exhibit high disagreement, encouraging exploration over exploitation to build sufficient confidence on the best model. 
In contrast, a lower $\epsilon$ increases the belief in the labeled examples when model disagreement is low.
In this regard, the effectiveness of tuning $\epsilon$ using noisy labels is intuitive: in our simplified single-parameter model, $\epsilon$ serves as a proxy for measuring disagreement among competing models regarding the data labels. By learning $\epsilon$ directly from these noisy labels generated by model predictions, we can accurately gauge the extend of it without requiring any labeled data. Although these noisy labels alone may not reliably identify the best or nearly best model (as shown in \Cref{sec:app_epsilon}), they still offer valuable insights that guide \framework{} in making more informed model selection decisions.

\section{Experiments}
\label{sec:experiments}
We conduct a comprehensive set of experiments to evaluate the performance of \framework{} for model selection. Our experiments span $8$ vision tasks and $8$ text classification tasks, each varying in number of classes. We test across 18 model collections with more than 1,500 pretrained models.

\subsection{Datasets and Model Collections}
For image classification tasks, we use $3$ ImageNet V2 datasets~\citep{recht2019imagenet} with $1,000$ classes and $10,000$ examples. We download more than $100$ pretrained models for each dataset from \citet{pytorchhub}, and further fine-tune them with different hyperparameters, resulting in varying accuracies. The model architectures range from ResNet~\citep{he2016deep} and MobileNet~\citep{howard2017mobilenetsefficientconvolutionalneural} to EfficientNet~\citep{tan2020efficientnetrethinkingmodelscaling}, with accuracy ranging from $43\%$ to $87\%$.

We also consider the PACS dataset~\citep{li2017deeperbroaderartierdomain}, in order to test model selection in the domain adaptation setting, which includes four domains. We train $30$ models using the architectures mentioned above, resulting in a range of model accuracy from $73\%$ to $94\%$.

For text classification tasks, we add $8$ datasets from the GLUE benchmark~\citep{wang2019gluemultitaskbenchmarkanalysis}, with test set sizes ranging from $71$ to $40,430$. We use pretrained models from \citet{huggingface} without further fine-tuning. The model collections include architectures such as BERT~\citep{devlin2019bertpretrainingdeepbidirectional}, DistilBERT~\citep{sanh2020distilbertdistilledversionbert}, and RoBERTa~\citep{liu2019robertarobustlyoptimizedbert}. The number of models in each collection varies from $80$ to $110$, and accuracy ranges from $5\%$ to $95\%$. The GLUE benchmark predicts $2$ or $3$ classes depending on the dataset.

Additionally, we use the same datasets and pretrained models as \citet{karimi2021online}, who study the online setting. The datasets we use include CIFAR-10~\citep{krizhevsky2009learning}, ImageNet~\citep{deng2009imagenet}, Drift~\citep{drift}, and EmoContext~\citep{semeval}.

We refer to  \Cref{sec:app_datasets} for additional details on datasets and model collections.

\subsection{Baselines}
\label{sec:baselines}
To evaluate \framework{} we compare against random sampling (\textsc{random}), as well as several adapted strategies listed below.

\begin{itemize} [itemsep=.5pt, topsep=0.25pt]
    \setlength{\leftskip}{-10pt} 
     \item \underline{Uncertainty Sampling.} We adapt the method of \citet{dagan1995committee} for model selection by creating a probability distribution over classes based on model predictions for each example. We then rank examples by their entropy and select the top $b$ example with the highest entropy. 
    \item \underline{Margin Sampling.} Adapted by \citet{seung1992query,freund1997selective}. The \textsc{margin} method is non-adaptive and selects examples with the smallest margin, defined as the difference between the highest and second-highest class probabilities.
    \item \underline{Active Model Comparison.} We adapt \textsc{amc} from \citet{NIPS2012_92fb0c6d}, which optimizes sampling to reduce the likelihood of selecting a worse model. While originally proposed for comparing two models within a labeling budget, we extend it to evaluate all pairs of pretrained models for best model selection.
    \item \underline{Variance Minimization Approach.} We implement \textsc{vma} from \citet{matsuura2023active}, which samples examples to minimize the variance of estimated risk. While \textsc{vma} requires softmax predictions, our setup treats models as black boxes, so we substitute softmax predictions with a probability distribution based on model predictions.
\end{itemize}
Notably, none of these methods are explicitly designed for pretrained model selection with limited labels, yet we adapt them as baseline comparisons within this newly formalized model selection framework.

\subsection{Experimental Setup}
\subsubsection{Evaluation Protocol and Learning $\epsilon$}\label{sec:eval-protocol}
We employ the following evaluation protocol. We uniformly sample $n$ i.i.d. instances from the entire test data. Each algorithm sequentially queries up to a labeling budget $b$ and selects the model with the highest accuracy on these $b$ labeled examples. We consider the model achieving the highest accuracy across all $n$ labels as the \emph{true} best model. We then evaluate each method by comparing its selected model based on $b$ labels to the true best model. We call this process \emph{realization} and repeat it multiple times to obtain a performance estimate for each method.

We learn the parameter $\epsilon$ for \framework{} directly from the model predictions, as mentioned in \Cref{sec:epsilon}. We first perform noisy labeling by assigning each data example the class most predicted by our models. For datasets with a large number of classes, we assign labels randomly based on the probability distributions formed by the model predictions. We then run our evaluation protocol treating our noisy labels as oracle labels and perform a grid search over different values of $\epsilon$. Finally, we select the $\epsilon$ that makes \framework{} perform best in identifying the best model, solely with the noisy labels.

Our experiments yield several insightful observations regarding the selected values of the parameter $\epsilon$. For a comprehensive analysis of these findings, we refer to~\Cref{sec:app_epsilon}. 

\subsubsection{Performance Metrics}
For a given labeling budget, we compare different baselines using the following key performance metrics:
\textit{Identification Probability}, defined as the fraction of realizations where a method successfully identifies the true best model of that realization, 
\textit{Label Efficiency}, the reduction in number of labels (\%) required to select the best or a near-best model over all realizations, specifically one within $\delta$ vicinity of the best model, and
\textit{$95th$ Percentile Accuracy Gap}, represents the $95$th percentile of the accuracy gap across all realizations, which is measured relative to the accuracy of the best model in each realization.

\subsection{Experimental Results}
We present our numerical results for each of the previously introduced performance metrics. Extended results can be found in \Cref{sec:app_extended_results}. Our observations for each metric are summarized as follows:

\subsubsection{Best Model Identification Probability}\label{sec:identification_prob}
\begin{figure*}[!h]
    \centering
    \includegraphics[width=0.9\linewidth]{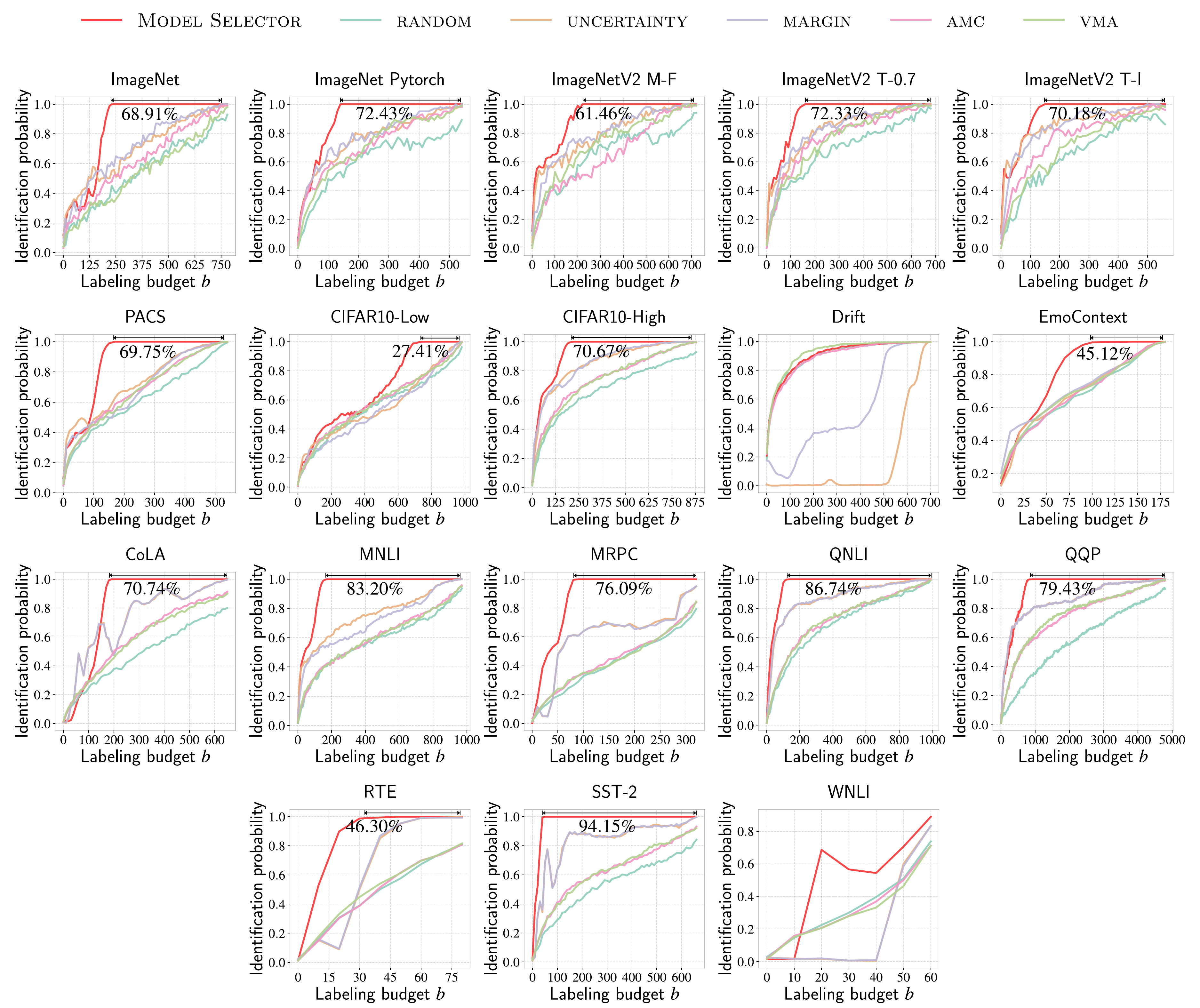} 
    \caption{Best model identification probability of \framework{} and the baselines on $18$ model collections. \framework{} is capable of reducing the labeling cost by up to $94.15\%$ for identifying the best model.} 
    \label{fig:succ_prob}
\end{figure*}

\Cref{fig:succ_prob} shows identification probabilities of \framework{} compared to baselines across 18 model collections, with labeling budgets extending until the best baseline (often \textsc{uncertainty}) reaches identification probability $100\%$ (see \Cref{sec:app_datasets} for details). \framework{} requires substantially fewer labels to reach a high identification probability than the baseline methods. For instance, our method reduces the labeling cost by up to $94.15\%$ to identify the true best model with $100\%$ identification probability, compared to the best competing baseline (primarily \textsc{uncertainty}). 
Compared to methods on ImageNet model collections, \framework{} reduces the labeling cost by $68.91\%, 69.35\%, 70.56\%, 71.56\%$, and $74.38\%$ for \textsc{margin}, \textsc{uncertainty}, \textsc{amc}, \textsc{vma}, and \textsc{random}, respectively.
We also examine a challenging scenario using the model collections on the Drift dataset, where there is a significant distribution shift between the training data for each model in the collection, and the model performs very differently on the unlabeled data pool. 
In such a setting, \textsc{random} is considered the strongest baseline compared to selective sampling methods~\citep{settles2009active,ayed2023data}.
By learning the parameter $\epsilon$ in a self-supervised manner from model predictions, \framework{} automatically detects the drift and adjusts $\epsilon$ to 0.5, effectively mimicking random sampling in these settings. 
Additionally, \framework{} outperforms existing baselines on the WNLI dataset using fewer than 70 examples, demonstrating its ability to identify the best model even on datasets with a limited number of labels.
Across different labeling cost, \framework{} can consistently match the identification probability of the baselines at a significantly lower labeling cost.
\begin{table}[!ht]
    \centering
\begin{tabular}{lccc}
\specialrule{1.5pt}{0pt}{0pt}
Dataset  & $\delta = 1\%$ & $\delta = 0.5\%$ & $\delta = 0.1\%$ \\
\midrule
CIFAR10-High & $\downarrow\textbf{48.04}\%$ & $\downarrow\textbf{58.40}\%$ & $\downarrow\textbf{72.23}\%$ \\
CIFAR10-Low & $\downarrow\textbf{21.07}\%$ & $\downarrow\textbf{21.82}\%$ & $\downarrow\textbf{25.67}\%$ \\
EmoContext & $\downarrow\textbf{20.56}\%$ & $\downarrow\textbf{34.19}\%$ & $\downarrow\textbf{39.89}\%$ \\
PACS & $\downarrow\textbf{62.73}\%$ & $\downarrow\textbf{66.81}\%$ & $\downarrow\textbf{68.62}\%$ \\
Drift & $\uparrow23.79\%$ & $\uparrow7.96\%$ & $\uparrow11.18\%$ \\
ImageNet & $\downarrow\textbf{53.62}\%$ & $\downarrow\textbf{63.80}\%$ & $\downarrow\textbf{69.81}\%$ \\
ImageNet Pytorch & $\downarrow\textbf{40.94}\%$ & $\downarrow\textbf{64.07}\%$ & $\downarrow\textbf{73.36}\%$ \\
ImageNetV2 T-I & $\uparrow6.12\%$ & $\downarrow\textbf{49.12}\%$ & $\downarrow\textbf{70.61}\%$ \\
ImageNetV2 T-0.7 & $\downarrow\textbf{57.58}\%$ & $\downarrow\textbf{57.79}\%$ & $\downarrow\textbf{73.39}\%$ \\
ImageNetV2 M-F & $\downarrow\textbf{48.18}\%$ & $\downarrow\textbf{61.39}\%$ & $\downarrow\textbf{56.72}\%$ \\
MRPC & $\downarrow\textbf{72.41}\%$ & $\downarrow\textbf{73.62}\%$ & $\downarrow\textbf{74.54}\%$ \\
CoLA & $\downarrow\textbf{45.89}\%$ & $\downarrow\textbf{53.75}\%$ & $\downarrow\textbf{71.01}\%$ \\
QNLI & $\downarrow\textbf{46.88}\%$ & $\downarrow\textbf{78.39}\%$ & $\downarrow\textbf{85.75}\%$ \\
QQP & $\uparrow11.90\%$ & $\downarrow\textbf{26.55}\%$ & $\downarrow\textbf{73.36}\%$ \\
SST-2 & $\downarrow\textbf{7.89}\%$ & $\downarrow\textbf{39.66}\%$ & $\downarrow\textbf{93.33}\%$ \\
WNLI & $0.00\%$ & $0.00\%$ & $0.00\%$ \\
MNLI & $\downarrow\textbf{69.42}\%$ & $\downarrow\textbf{79.83}\%$ & $\uparrow3.95\%$ \\
RTE & $\downarrow\textbf{40.96}\%$ & $\downarrow\textbf{40.96}\%$ & $\downarrow\textbf{40.96}\%$ \\
\specialrule{1.5pt}{0pt}{0pt}
\end{tabular}
    \caption{Label efficiency for near-best models: \framework{} consistently reduces labeling cost to reach the $\delta$ vicinity of the true best model compared to the best competing method.}
   
    \label{tab:label_eff}
\end{table}

\subsubsection{Label Efficiency for Near-Best Models}\label{sec:near_best}
We examine labeling efficiency from a different perspective. We evaluate the budget required to select the best \emph{or} a near-best model with accuracy within the $\delta$ vicinity of that of the best model. Specifically, we measure the required number of labels where, in all realizations, the selected models are within $1\%$, $0.5\%$, and $0.1\%$ of the best model accuracy. 

\Cref{tab:label_eff} shows the percentage reduction in the number of labels required by \framework{} to select a near-best model over all realizations.
To reach the same $\delta$ vicinity of the accuracy of the best model, our method requires fewer labels than the best competing baselines (mainly \textsc{uncertainty} and \textsc{margin}). 
Quantitatively, for MRPC, \framework{} reduces the labeling cost by $72.41\%$, $73.62\%$, and $74.54\%$ for $\delta$ values of $1\%$, $0.5\%$ and $0.1\%$, respectively. 
Moreover, our method approaches the performance of the near-best method on ImageNetV2 for all model collections. Specifically, \framework{} with $\delta=0.1\%$ reduces the labeling cost by $70.61\%, 73.39\%$, and $56.72\%$ for ImageNetV2 T-I, ImageNetV2 T-0.7, and ImageNetV2 M-F, respectively. 
Our results show that \framework{} is consistently more label-efficient, not only for identifying the best model but also for selecting near-best models across different settings.

\begin{table*}[!ht]
    \centering
    \resizebox{1\textwidth}{!}{%
\begin{tabular}{l c c c c c c}
\specialrule{1.5pt}{0pt}{0pt}
Dataset & \framework{} & \textsc{random} & \textsc{margin} & \textsc{uncertainty} & \textsc{amc} & \textsc{vma} \\
Identification probability & \small(70\%/80\%/90\%/100\%) & \small(70\%/80\%/90\%/100\%) & \small(70\%/80\%/90\%/100\%) & \small(70\%/80\%/90\%/100\%) & \small(70\%/80\%/90\%/100\%) & \small(70\%/80\%/90\%/100\%) \\
\midrule
CIFAR10-High & \textbf{1.90/0.80/0.40/0.00} & 5.00/3.90/3.50/3.00 & \underline{2.00}/\underline{1.30}/1.30/1.10 & 2.50/1.50/\underline{1.00}/\underline{0.70} & 3.80/2.60/2.10/1.80 & 4.00/3.00/2.60/1.90 \\
CIFAR10-Low & \textbf{1.40/0.90/0.50/0.00} & 2.00/1.80/1.40/1.30 & 2.10/1.80/1.60/1.40 & 2.10/1.80/1.50/1.30 & \underline{1.70}/1.40/\underline{1.20}/\underline{1.10} & 2.00/\underline{1.60}/1.50/1.30 \\
EmoContext & 1.30/\textbf{0.60/0.30/0.00} & \textbf{1.10}/\underline{1.00}/0.90/0.70 & 2.00/\underline{1.00}/0.80/\underline{0.50} & 1.50/1.10/\underline{0.70}/\underline{0.50} & 1.40/\underline{1.00}/0.80/\underline{0.50} & \underline{1.20}/\underline{1.00}/0.90/\underline{0.50} \\
PACS & \textbf{1.40/1.10/0.40/0.00} & 1.90/1.70/1.70/1.70 & 1.80/1.80/1.80/1.80 & \underline{1.70}/\underline{1.60}/\underline{1.50}/1.50 & \underline{1.70}/\underline{1.60}/\underline{1.50}/\underline{1.40} & 1.80/\underline{1.60}/1.70/1.50 \\
Drift & \textbf{11.33}/8.27/\underline{5.87}/\textbf{0.00} & \underline{11.47}/\textbf{7.87}/6.27/\textbf{0.00} & 16.67/16.67/13.87/\underline{7.60} & 18.00/17.33/10.00/10.00 & 11.87/\underline{8.13}/6.53/\textbf{0.00} & 11.60/9.47/\textbf{3.60}/\textbf{0.00} \\
ImageNet & \textbf{0.90/0.90/0.80/0.00} & 2.30/2.20/2.10/2.10 & 1.20/\underline{1.20}/\underline{1.20}/\underline{1.10} & \underline{1.10}/1.50/1.30/1.30 & 1.70/1.70/1.30/1.40 & 1.70/1.70/1.70/1.70 \\
ImageNet Pytorch & \textbf{0.80/0.50/0.20/0.00} & 3.70/3.30/3.00/2.60 & 1.30/\underline{0.90}/\underline{0.80}/\underline{0.70} & \underline{1.00}/1.00/1.00/0.80 & 2.20/1.90/1.30/1.20 & 3.60/2.40/1.90/1.20 \\
ImageNetV2 T-I & \textbf{1.20/0.70/0.10/0.00} & 4.30/4.50/3.00/2.20 & \underline{1.30}/1.30/1.10/\underline{0.50} & 1.70/1.40/\underline{0.60}/0.60 & 3.50/2.80/1.90/1.80 & 3.10/2.40/2.30/1.60 \\
ImageNetV2 T-0.7 & \textbf{1.00/0.50/0.20/0.00} & 4.20/3.70/3.50/2.50 & \underline{1.50}/\underline{1.30}/\underline{1.10}/1.10 & \underline{1.50}/1.50/1.30/\underline{1.00} & 2.60/2.40/1.80/1.30 & 2.80/2.70/2.30/1.80 \\
ImageNetV2 M-F & \textbf{0.90/0.40/0.30/0.00} & 4.10/2.60/2.60/2.60 & \underline{1.10}/\underline{1.00}/0.90/0.60 & \underline{1.10}/1.10/0.90/0.60 & 3.10/1.10/1.10/0.90 & 3.70/1.70/1.60/1.60 \\
MRPC & \textbf{1.14/1.14/0.29/0.00} & 5.71/5.43/5.14/4.86 & 2.00/\underline{1.43}/1.14/\underline{0.86} & \underline{1.71}/\textbf{1.14}/\underline{0.86}/1.14 & 5.43/5.14/5.14/4.86 & 5.43/5.14/4.86/4.29 \\
CoLA & \textbf{0.88/0.62/0.25/0.00} & 3.38/3.37/3.37/3.12 & 1.12/\underline{0.88}/\underline{1.12}/\underline{1.37} & \underline{1.00}/\underline{0.88}/\underline{1.12}/\underline{1.37} & 2.62/2.50/2.50/2.37 & 2.50/2.50/2.38/2.25 \\
QNLI & \textbf{1.00/0.60/0.30/0.00} & 4.60/4.20/3.90/3.80 & \underline{2.10}/\underline{1.40}/\underline{1.00}/\underline{0.80} & 2.40/\underline{1.40}/\underline{1.00}/\underline{0.80} & 4.60/4.00/3.80/3.60 & 4.40/4.20/3.90/3.60 \\
QQP & 0.46/\textbf{0.24/0.12/0.00} & 1.50/1.44/1.36/1.30 & \underline{0.40}/\underline{0.30}/\underline{0.26}/\underline{0.22} & \textbf{0.38}/\underline{0.30}/\underline{0.26}/0.24 & 1.08/0.96/0.80/0.72 & 1.10/0.90/0.80/0.72 \\
SST-2 & \textbf{0.40/0.27/0.13/0.00} & 6.80/6.40/6.27/6.40 & \underline{0.40}/0.40/0.40/0.40 & \underline{0.40}/0.40/0.40/0.40 & 5.87/5.60/5.60/5.47 & 5.73/5.47/5.33/5.33 \\
WNLI & \textbf{3.08/3.08/1.54/0.00} & 12.31/\underline{4.62}/\textbf{1.54}/\underline{1.54} & \underline{6.15}/\textbf{3.08}/\textbf{1.54}/\underline{1.54} & \underline{6.15}/\textbf{3.08}/\textbf{1.54}/\underline{1.54} & 9.23/\textbf{3.08}/\underline{3.08}/\underline{1.54} & 9.23/\textbf{3.08}/\underline{3.08}/\underline{1.54} \\
MNLI & \textbf{1.00/0.80/0.40/0.00} & 4.70/4.30/3.90/2.90 & 1.20/1.20/\underline{1.10}/\underline{1.00} & \underline{1.10}/\underline{1.10}/\underline{1.10}/\underline{1.00} & 4.40/4.00/3.20/2.70 & 4.00/4.30/3.70/2.90 \\
RTE & \textbf{10.40/10.00/4.40/0.00} & 22.40/21.20/16.40/11.20 & 21.20/\underline{16.80}/17.20/\underline{6.80} & \underline{20.80}/\underline{16.80}/19.20/7.20 & 22.80/20.80/16.80/10.80 & 22.00/19.20/\underline{14.80}/10.80 \\
\specialrule{1.5pt}{0pt}{0pt}
\end{tabular}
    }
    \caption{Robustness analysis: $95$-th Percentile Accuracy Gap ($\%$) at budget needed for \framework{} to reach identification probability $70\%, 80\%, 90\%$, and $100\%$. Compared to baselines, \framework{} achieves a smaller accuracy gap from the best model.
    Best method bolded; Next best underlined.}
    \label{tab:percentile_combined}
\end{table*}

\subsubsection{Robustness Analysis}\label{sec:robustness}
We compute the $95$-th percentile of the accuracy gap to assess the robustness of \framework{} and perform worst case scenario analysis for each method.
Specifically, we calculate the accuracy gap between the selected model and the true best model for all 
realizations and select the accuracy gap that is larger than $95\%$ of accuracy gaps across all 
realizations.
We evaluate this for different budgets for each
dataset, determined as the budget required
for \framework{} to achieve $70\%$, $80\%, 90\%$, and $100\%$
identification probability. 

As shown in \Cref{tab:percentile_combined}, \framework{} achieves significantly smaller accuracy gaps compared to baseline methods.
For example, the best competing methods (\textsc{margin} and \textsc{uncertainty}) on the RTE dataset with high disagreement among the model predictions,
for identification probability of $70\%, 80\%$, and $90\%$, return a model with $20.80\%$, $16.80\%$, and $17.20\%$ accuracy gaps, while \framework{} returns models with accuracy gaps that are $10.40\%, 10.00\%$, and $0.40\%$. 
Quantitatively, these are $2\times,  1.7\times$, and $43\times$ smaller accuracy gaps.  
Compared to each method for MNLI at identification probability $90\%$, \framework{} selects the model with $0.4\%$ accuracy gap, while \textsc{margin}, \textsc{uncertainty}, \textsc{amc}, \textsc{vma}, and \textsc{random} select models with accuracy gaps of
$1.1\%$, $1.1\%$, $3.2\%$, $3.7\%$, and $3.9\%$, which is $2.8\times$, $2.8\times$, $8\times$, and $9.8\times$ larger compared to our method. 
Even in 
settings with a small number of examples, such as WNLI, and datasets with a large number of classes, such as ImageNet, \framework{} outperforms the baselines. Specifically, when \framework{} reaches an identification probability of $100\%$, the best competing baselines still select models with accuracy gaps of $1\%$ and $1.1\%$ for WNLI and ImageNet, respectively.

Our findings highlight the robustness of \framework{} for consistently returning a near-best model even at its lowest performance.

\section{Discussions}
\label{sec:discussion}
\framework{} delivers competitive performance across various settings by capturing the discrepancy between the  candidate models and the true labels with just a single parameter $\epsilon$. Additionally, it relies solely on hard predictions, without requiring access to the internal workings of pretrained models. Our future work will extend to settings where soft predictions are available and explore model selection with limited demonstrations for generative models.

\section*{Acknowledgements}
We would like to thank Ce Zhang, Maurice Weber, Luka Rimanic, Jan van Gemert and David Tax for their input on this work.
This research is carried out in the frame of the “UrbanTwin:
An urban digital twin for climate action; Assessing policies
and solutions for energy, water and infrastructure” project
with the financial support of the ETH-Domain Joint Initiative
program in the Strategic Area Energy, Climate and Sustainable
Environment. We would like to thank the Swiss National
Supercomputing Centre (CSCS) for access and support of the
computational resources.

\bibliography{arxiv}
\bibliographystyle{arxiv}

\appendix
\onecolumn
\section*{\centering \LARGE \bfseries Appendix}
\noindent\rule{\textwidth}{0.5pt}  
\vspace{0mm}
\section{Datasets and Model Collections}
\label{sec:app_datasets}

\Cref{tab:data_info} summarizes the details of the $18$ model collections used in \Cref{sec:experiments}. These collections vary in dataset size, realization pool size, range of pretrained model accuracies, number of pretrained models, and number of classes.
The dataset sizes vary from as few as $71$ examples for WNLI to $50,000$ for ImageNet.
The number of predicted classes ranges from binary classification tasks in the GLUE benchmark to $1,000$ classes for ImageNet.
We have experimented with settings involving as few as $8$ models for EmoContext up to $118$ models for WNLI.
We choose the realization pool size to be practical from a practitioner's point of view; specifically, we avoid labeling the entire test dataset. For most datasets, the realization pool size is $1,000$. For model collections on ImageNet and ImageNetV2, we do not use larger realization pool sizes to enable comparison with baseline methods, which otherwise would cause out-of-memory exceptions and have long execution times (mostly \textsc{vma}).
For every model collection in \Cref{sec:experiments} we evaluate for all budgets $b$ up to the size of the realization pool.

In \Cref{tab:data_info} we also show the best found $\epsilon$ parameter across $1,000$ realizations for every model collection. 
Further details on selection of $\epsilon$ can be found in \Cref{sec:app_epsilon}.

In conclusion, \Cref{tab:data_info} highlights the diverse range of settings considered in our experiments.

\begin{table}[htbp]
    \centering
\begin{tabular}{lccccccc}
\specialrule{1.5pt}{0pt}{0pt}
Dataset & Best $\epsilon$ across & Dataset  & Realization & Model & Number of & Number of \\
 & 1,000 realizations & size & pool size & accuracy & models & classes \\
\midrule

CIFAR10-High & 0.47 & 10,000 & 1,000  & 55\% - 92\% & 80 & 10 \\
CIFAR10-Low  & 0.47 & 10,000 & 1,000 & 40\% - 70\% & 80 & 10 \\
EmoContext   & 0.47 & 5509 & 1,000 & 88\% - 92\%  & 8 & 4 \\
PACS         & 0.45 & 9991 & 1,000 & 73\% - 94\%  & 30 & 7 \\
Drift        & 0.50 & 3600 & 1,000 & 25\% - 60\%  & 9 & 6 \\
ImageNet     & 0.45 & 50,000 & 1,000 & 50\% - 82\%  & 102  & 1,000 \\
ImageNet Pytorch & 0.45 & 50,000 & 1,000 & 55\% - 87\%  & 114 & 1,000 \\
ImageNetV2 T-I & 0.46 & 10,000 & 1,000 & 58\% - 89\%  & 114 & 1,000 \\
ImageNetV2 T-0.7 & 0.45 & 10,000 & 1,000 & 51\% - 86\%  & 114  & 1,000 \\
ImageNetV2 M-F & 0.48 & 10,000 & 1,000 & 43\% - 81\%  & 114 & 1,000\\
MRPC         & 0.37 & 408 & 350  & 31\% - 91\%  & 95 &  2 \\
CoLA         & 0.45 & 1043 & 800 & 14\% - 87\%  & 109 & 2 \\
QNLI         & 0.44 & 5463 & 1,000 & 16\% - 90\%  & 90 & 2 \\
QQP          & 0.47 & 40430 & 5,000 & 8\% - 85\%  & 101 & 2 \\
SST-2        & 0.36 & 872 & 750 & 7\% - 97\%  & 97 & 2 \\
WNLI         & 0.47 & 71 & 65 & 8\% - 59\%  & 118 & 2 \\
MNLI         & 0.43 & 9815 & 1,000 & 5\% - 91\%  & 82 & 3 \\
RTE          & 0.39 & 277 & 250 & 41\% - 88\%  & 87 & 2 \\
\specialrule{1.5pt}{0pt}{0pt}
\end{tabular}
    \caption{Summary of the $18$ model collections and datasets used in our experiments, including dataset sizes, realization pool sizes, ranges of pretrained model accuracies, numbers of pretrained models, and numbers of classes.}
    \label{tab:data_info}
\end{table}

\begin{figure*}[hbtp]
    \centering
    \includegraphics[width=\linewidth]{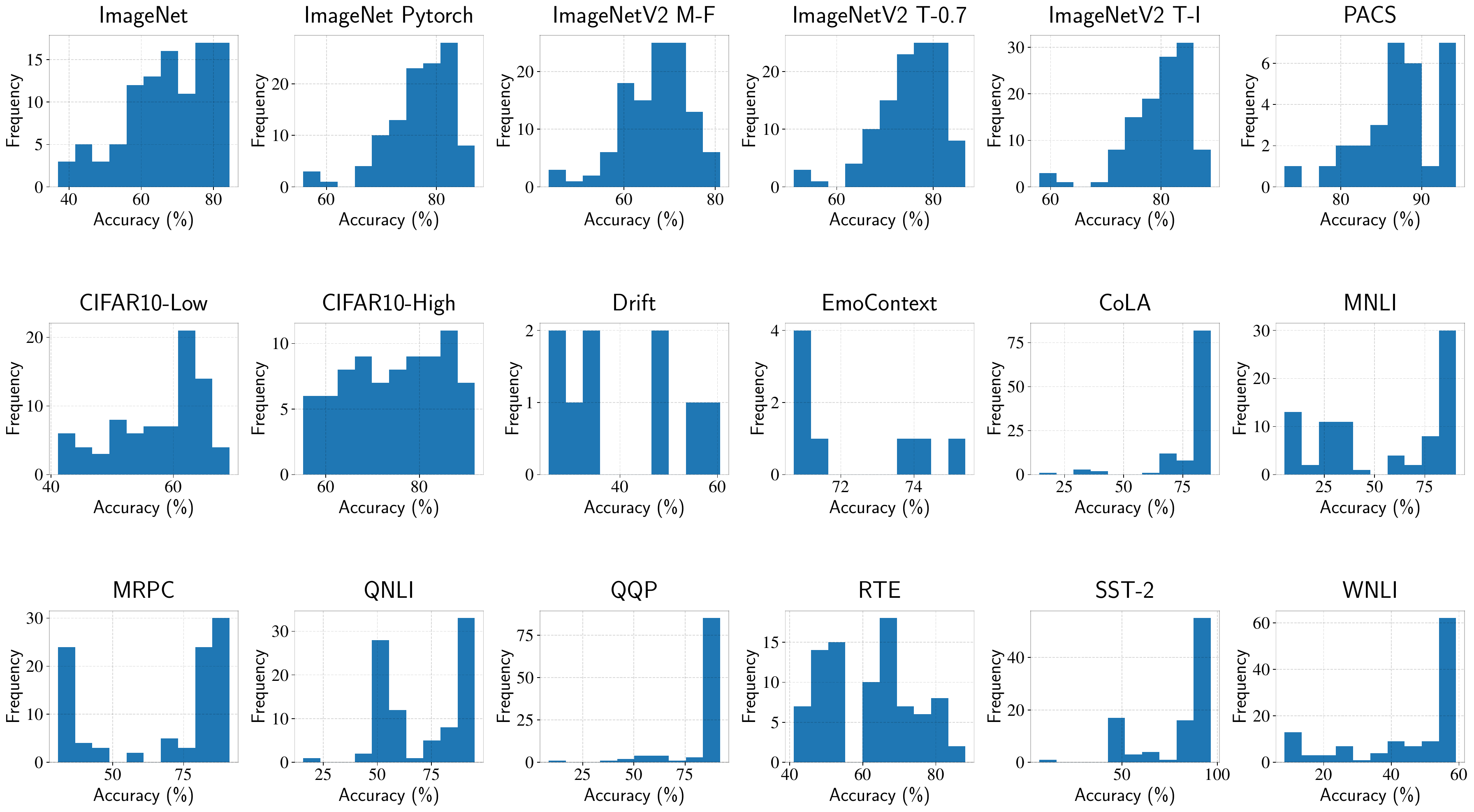} 
    \caption{Accuracies of the models used in \Cref{sec:experiments}, evaluated on the entire dataset. Our experiments cover different scenarios across a wide range of model accuracies.} 
    \label{fig:model_accuracies}
\end{figure*}

\Cref{fig:model_accuracies} displays the accuracies of the models used in \Cref{sec:experiments}, evaluated on the entire dataset.
Our experiments encompass model collections ranging from those where the majority of models perform equally well, such as QQP and CoLA, to settings where there is an equal distribution of high-performing and low-performing models, like CIFAR10 and ImageNet. We also consider settings where most models do not perform well and have high accuracy, such as Drift and RTE.

This demonstrates that we capture a wide range of possible model accuracies without making any assumptions about their distribution.

\section{Baselines}
\label{sec:baselines}
To evaluate \framework{} we compare against random sampling (\textsc{random}), as well as uncertainty sampling (\textsc{uncertainty})~\citep{dagan1995committee} and margin sampling (\textsc{margin})~\citep{seung1992query,freund1997selective}. Additionally, we use the variance minimization approach (\textsc{vma})~\citep{matsuura2023active} and active model comparison (\textsc{amc})~\citep{NIPS2012_92fb0c6d}.

\paragraph{Uncertainty Sampling.} We adapt the method of \citet{dagan1995committee} to our context. For each example, we create a probability distribution over the possible classes by counting predictions of all models. We sort the examples according to the probability distribution with maximal uncertainty, that is, entropy. Then we select $b$ examples with highest entropy. Note, that this is a non-adaptive baseline, meaning after querying a label it does not change the decision for the next example.

\paragraph{Margin Sampling.} In a similar manner to \textsc{uncertainty}, we adapt the method for margin sampling, as proposed by \citet{seung1992query,freund1997selective}. The \textsc{margin} method selects examples with the largest margin, where margin is defined as the difference between the highest probability of a certain class and the second-highest probability of a different class. As for \textsc{uncertainty}, it selects $b$ examples with largest margin in a non-adaptive manner. 

\paragraph{Active Model Comparison.} We implement the Active Comparison of Predictive Models method proposed by \citet{NIPS2012_92fb0c6d}. \textsc{amc} samples from a distribution that maximizes the power of a statistical test, and with that minimizes the likelihood of selecting a worse model. While \textsc{amc} is originally designed to compare the risks of two predictive models within a fixed labeling budget, we extend the approach to evaluate all possible pairs of pretrained models, allowing for the selection of the best model within the given budget. 

\paragraph{Variance Minimization Approach.} We implement the algorithm proposed by \citet{matsuura2023active}. \textsc{vma} samples examples such that the variance of the estimated risk is minimized.
Although \textsc{vma} is designed for active model selection, it assumes to have a test loss estimator. Specifcally, it requires softmax predictions for each example. However, in our setup models are treated as black boxes. Therefore, instead of softmax predictions we use probability distribution over the possible classes from counting predictions of all models.

It is important to note that none of these methods are specifically tailored for our setting of pretrained model selection with limited labels. Nonetheless, we adapt these methods as baseline comparisons within this newly formalized model selection framework.

\section{Related Work}
\label{sec:app_rel_work}

\paragraph{Optimal Information Gathering}  
How to gather information optimally has been studied from different angles and for different settings.
The majority of works do not consider the model selection setting. For example, information gathering has been explored for active learning~\citep{mackay1992information,dasgupta2011active,ding2023learning,ding2024learning}, experimental design~\citep{lindley1956measure,fedorov2013theory}, reinforcement learning~\citep{treven2023optimistic}, evaluation of (stochastic) Boolean functions~\citep{kaplan2005learning,deshpande2014approximation}, active hypothesis testing~\citep{chernoff1992sequential,nowak2009noisy,naghshvar2013active}, channel coding with feedback~\citep{horstein1966sequential,burnashev1976data}, and feature selection~\citep{covert2023learning}. 
Furthermore, \citet{hübotter2024transductiveactivelearningtheory} discuss how to optimally gather information most efficiently. ~\citet{smith2018understanding} explore modes of failure for measuring uncertainty in adversarial settings.
~\citet{emmenegger2024likelihood} discuss gathering information in the sequential setting as we do, yet not specifically for model selection.
In the context of minimizing epistemic uncertainty about an unknown target, several methods utilize the most informative selection policy using mutual information for active learning setting~\citep{gal2017deep, houlsby2011bayesian, nguyen2021information, hubotter2024information,kirsch2024advancing}. Our work falls into this category. In particular, our motivation is to exploit the enjoyable properties of the most informative selection policy in the binary symmetric channel~\citep{chen2015sequential} for model selection under the uncertainty of the channel, where we consider the channel model as our epistemic uncertainty model. 
The provable near-optimality guarantees therein generally apply to our case. In instances not, our learned $\epsilon$ would be highest (simply $0.5$), indicating that simple random sampling should be the most competitive baseline.
Furthermore, works can be categorized based on the assumptions of underlying noise into the noise-free settings~\citep{freund1997selective,kosaraju1999optimal,dasgupta2004analysis,chakaravarthy2007decision,golovin2011adaptive} and works with noisy observations~\citep{balcan2006agnostic,hanneke2007bound,gonen2013efficient,balcan2007margin,tsybakov2004optimal,hanneke2015minimax}. Our work falls into the latter category.
Many works have analyzed active learning with noisy observations, similarly to our setting with noisy prediction estimates for the pool-based setting. Yet, they limit their theoretical insights to limited hypotheses classess~\citep{balcan2007margin,gonen2013efficient}, and restricted noise settings~\citep{tsybakov2004optimal,hanneke2015minimax}.
In all of the previous studies, mutual information repeats itself as a key concept for both label-efficient model selection and optimal information gathering. ~\citet{chattopadhyayperformance} provide theoretical bounds on using greedy mutual information, and others show it in practice~\citep{treven2023optimistic,chen2015sequential}.
In conclusion, our work falls into the category of works for sequential pool-based noisy information gathering by maximizing mutual information with provable near-optimality guarantees.

\section{The Parameter $\epsilon$}
\label{sec:app_epsilon}
In this chapter we present the results for choosing the $\epsilon$ parameter. 
We select the best $\epsilon$ as described in \Cref{sec:epsilon}, that is, by constructing the noisy oracle and observing the identification probability for different $\epsilon$ values.

\subsection{Selecting the Parameter $\epsilon$} 
\begin{figure*}[!htpb]
    \centering
    \vspace{-5mm}
    \includegraphics[width=0.9\linewidth]{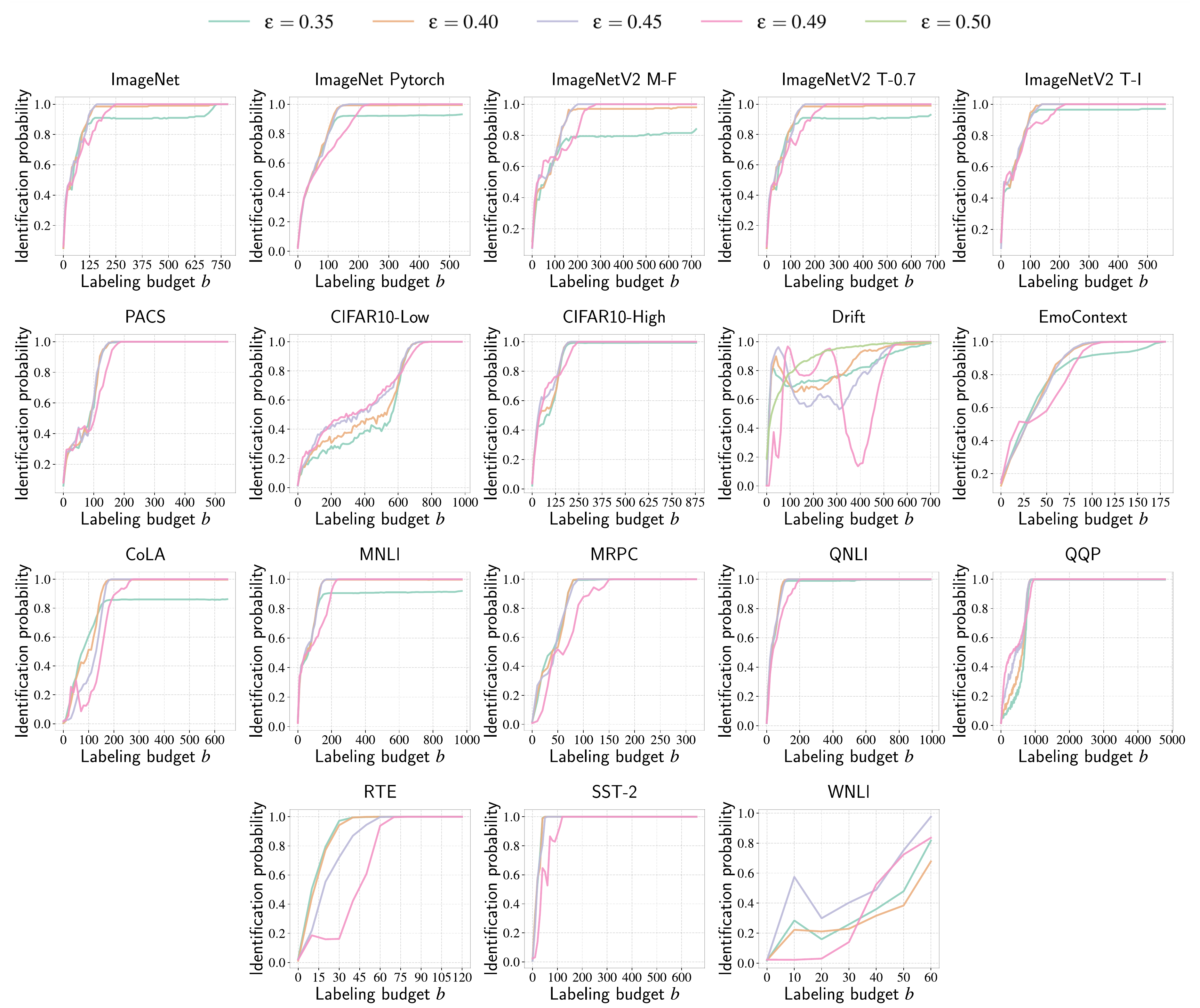} 
    \caption{Best model identification probability of \framework{} for $\epsilon \in \{0.35, 0.40, 0.45, 0.49, 0.50\}$ on $18$ model collections using oracle labels.} 
    \label{fig:gt_epsilon_coarse}
\end{figure*}

\begin{figure*}[!htbp]
    \centering
    \includegraphics[width=0.9\linewidth]{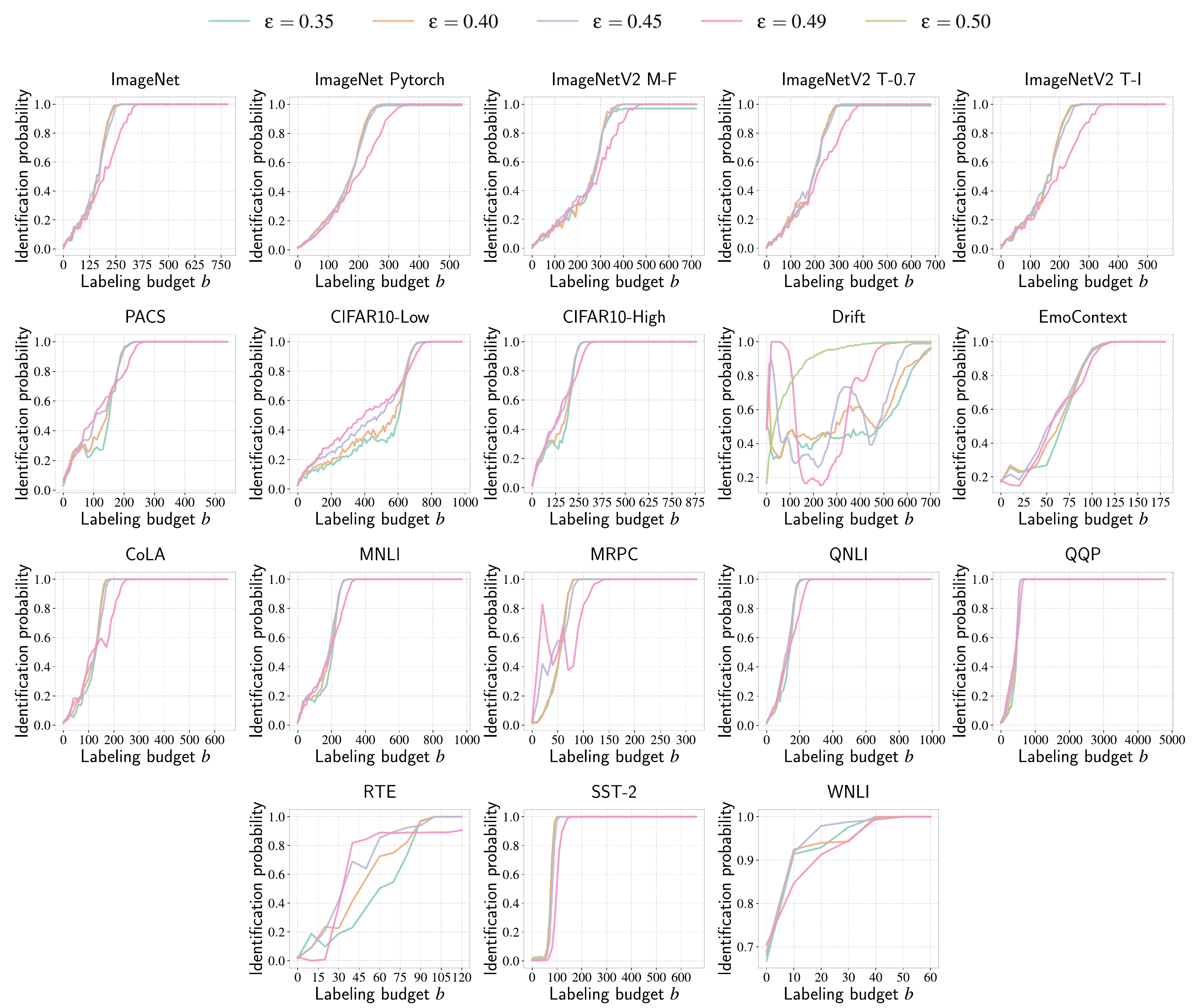} 
    \caption{Best model identification probability of \framework{} for $\epsilon \in \{0.35, 0.40, 0.45, 0.49, 0.50\}$ on $18$ model collections using noisy labels.} 
    \label{fig:nois_epsilon_coarse}
\end{figure*}

In \Cref{fig:nois_epsilon_coarse} we show the identification probability (see \Cref{sec:identification_prob}) of \framework{} with different $\epsilon$ values on 18 model collections using the noisy labels.
For comparison, in \Cref{fig:gt_epsilon_coarse} we show the same ranges of $\epsilon$ using the oracle labels. Both figures do not visualize the entire realization, and are cut at budgets for demonstrations.
\Cref{fig:gt_epsilon_coarse,fig:nois_epsilon_coarse} confirm our statement that appropriate values for $\epsilon$ lie in range $[0.35, 0.49]$. Smaller values than $\epsilon=0.35$ lead to overfitting (see \Cref{sec:epsilon}).
When comparing model collections that use noisy labels with those that use oracle labels, we find that, fewer labels are needed to achieve a high identification probability at the same $\epsilon$ values.
For example, $\epsilon=0.45$ requires fewer labels to reach high identification probability than $\epsilon=0.35$ for CIFAR10-Low in both \Cref{fig:nois_epsilon_coarse}  and  \Cref{fig:gt_epsilon_coarse}. Also, we observe that for all ImageNet model collections \framework{} with $\epsilon=0.49$ requires more labels to reach high identification probabilities when compared to smaller $\epsilon$ values using both noisy and oracle labels.
In addition, this experiment demonstrates the robustness of \framework{} to different $\epsilon$ values. Quantitatively, QQP and QNLI demonstrate that all observed $\epsilon$ values ($0.35, 0.40, 0.45, 0.49$) achieve high identification probability requiring approximately equal amount of labels from the entire realization.

\begin{figure*}[!htbp]
    \centering
    \includegraphics[width=0.9\linewidth]{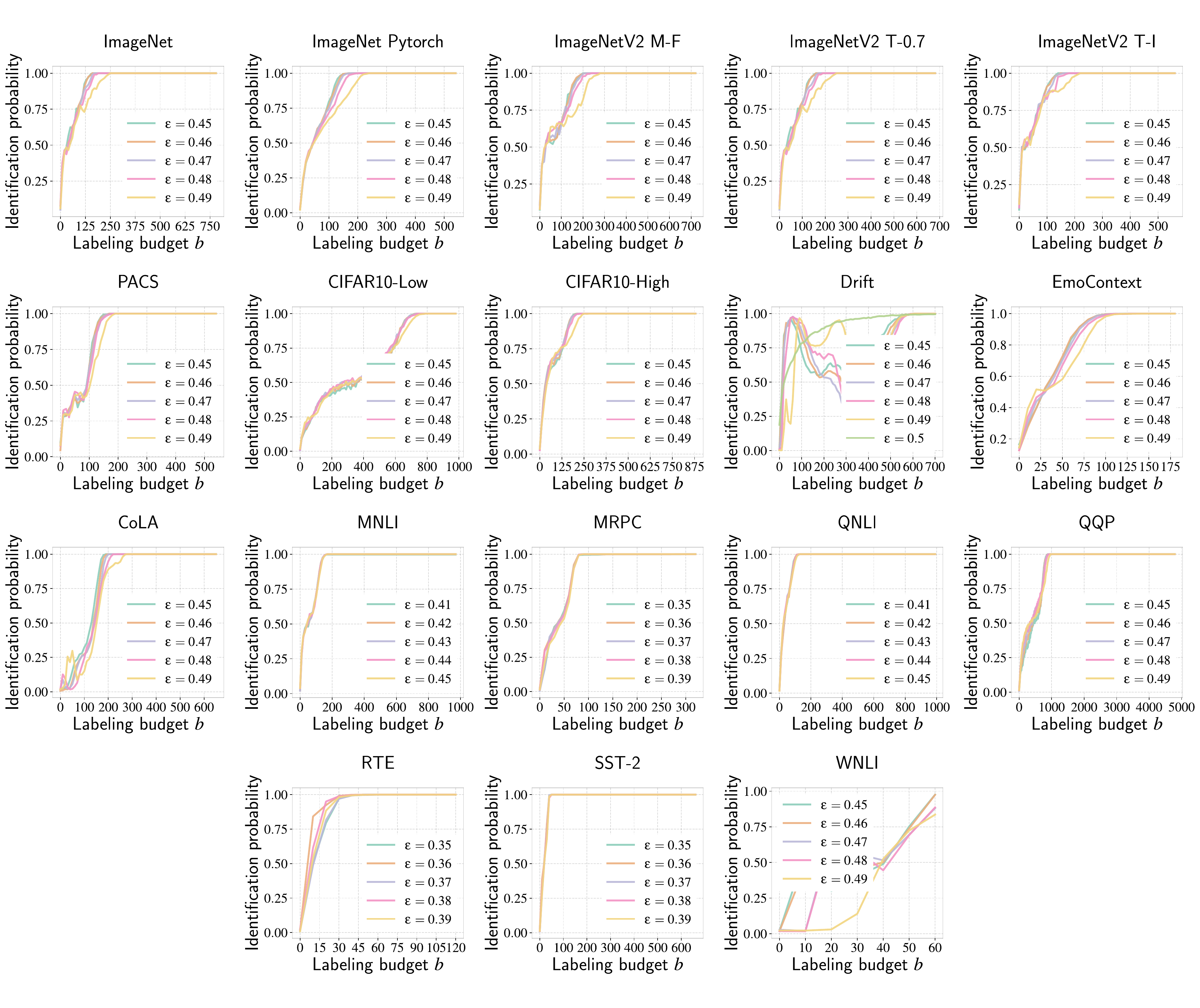} 
    \caption{Best model identification probability of \framework{} for finely grained $\epsilon$ values on $18$ model collections using oracle labels.} 
    \label{fig:gt_epsilon_fine}
\end{figure*}

\begin{figure*}[!hbtp]
    \centering
    \includegraphics[width=0.9\linewidth]{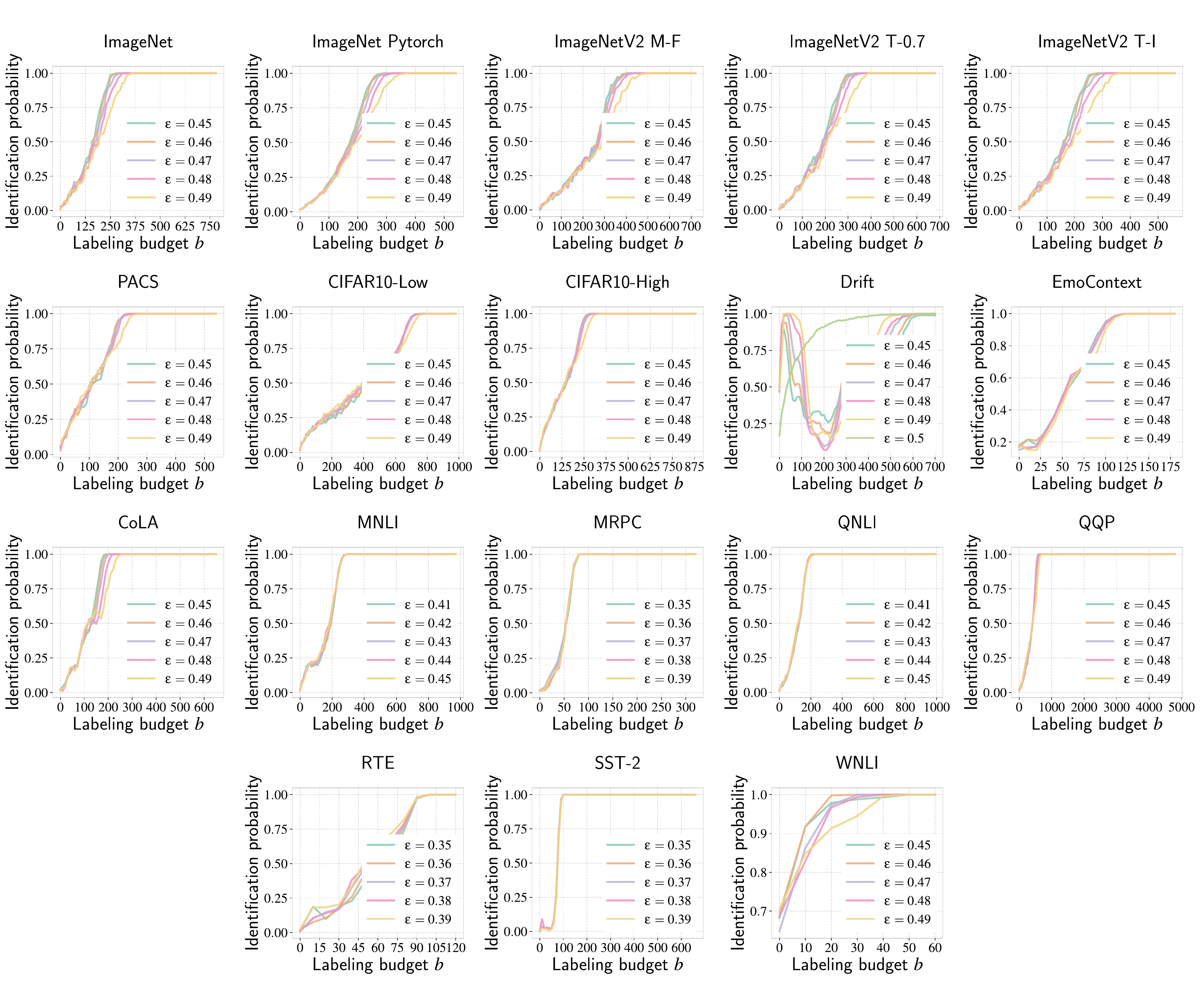} 
    \caption{Best model identification probability of \framework{} for finely grained $\epsilon$ values on $18$ model collections using noisy labels.} 
    \label{fig:nois_epsilon_fine}
\end{figure*}

Having identified the $\epsilon$ value ranges that reduce the labeling cost required to achieve the same identification probability, we perform the same experiment using more finely grained $\epsilon$ ranges.
In \Cref{fig:gt_epsilon_fine} and \Cref{fig:nois_epsilon_fine} we show the identification probability of \framework{} with finely grained $\epsilon$ values on 18 model collections using the noisy labels and oracle labels, respectively.
\Cref{fig:gt_epsilon_fine,fig:nois_epsilon_fine} further show the robustness of \framework{} to $\epsilon$ values. Both from noisy labels and oracle labels one can see that \framework{} with $\epsilon \in [0.45, 0.49]$ reaches high identification probability with approximately the same amount of required labels for both CIFAR10 model collections, CoLA, and QQP. The same holds for  $\epsilon \in [0.41, 0.45]$ on MNLI and QNLI with both noisy and oracle labels.

As discussed in \Cref{sec:experiments} for Drift model collection, \framework{} automatically adjusts
$\epsilon$ to $0.5$, effectively mimicking random sampling. \framework{} is capable detecting that using only noisy labels.

When the figures do not clearly indicate which $\epsilon$ value has the highest reduction in labeling cost, we calculate it numerically to find the best $\epsilon$ values. Selected best $\epsilon$ values for every model collection using noisy labels and oracle labels can be found in \Cref{tab:best_epsilon}. Quantitatively, our estimation of $\epsilon$ has an error
margin of only $0.01$ compared to the values obtained
using the oracle labels.

\begin{table}[!htbp]
    \centering
    \vspace{-0mm}
\begin{tabular}{lcc}
\specialrule{1.5pt}{0pt}{0pt}
Dataset & $\epsilon$ on noisy labels & $\epsilon$ on oracle labels \\
\midrule

CIFAR10-High & 0.47 & 0.47 \\
CIFAR10-Low  &  0.47 & 0.47 \\
EmoContext   & 0.47 & 0.46 \\
PACS         & 0.45 & 0.45 \\
Drift        & 0.50 & 0.50 \\
ImageNet     & 0.45 & 0.46 \\
ImageNet Pytorch & 0.45 & 0.46 \\
ImageNetV2 T-I & 0.46 & 0.45 \\
ImageNetV2 T-0.7 & 0.45 & 0.45 \\
ImageNetV2 M-F & 0.48 & 0.47 \\
MRPC         & 0.37 & 0.36 \\
CoLA         & 0.45 & 0.45 \\
QNLI         & 0.44 & 0.45 \\
QQP          & 0.47 & 0.47 \\
SST-2        & 0.36 & 0.36 \\
WNLI         & 0.47 & 0.47 \\
MNLI         & 0.43 & 0.44 \\
RTE          & 0.39 & 0.39 \\
\specialrule{1.5pt}{0pt}{0pt}
\end{tabular}
    \caption{Selected $\epsilon$ using noisy labels compared to selected $\epsilon$ using oracle labels. Our estimation of $\epsilon$ has an error margin of only 0.01 compared to the values obtained using the ground truth oracle.}
    \label{tab:best_epsilon}
\end{table}

\subsection{Choosing $\epsilon$ with Initial Oracle Labels}
\begin{figure*}[!htbp]
    \centering
    \vspace{-5mm}
    \includegraphics[scale=0.4]{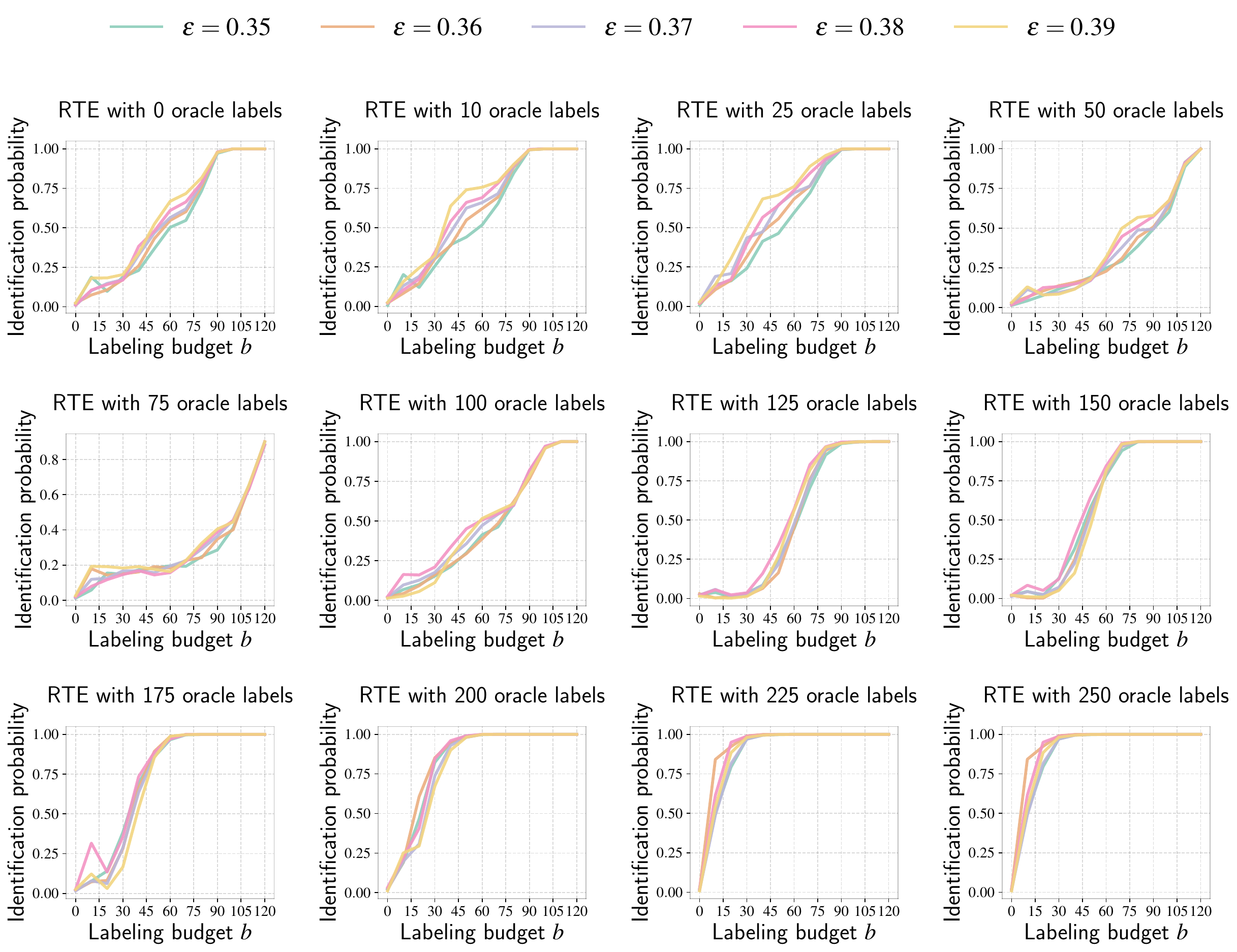} 
    \caption{Best model identification probability of \framework{} for $\epsilon \in [0.35, 39]$ on RTE dataset, as the number of oracle labels increases.} 
    \label{fig:noisy_to_gt_rte}
    \vspace{-3mm}
\end{figure*}

\begin{figure*}[!htbp]
    \centering
    \vspace{-0mm}
    \includegraphics[scale=0.36]{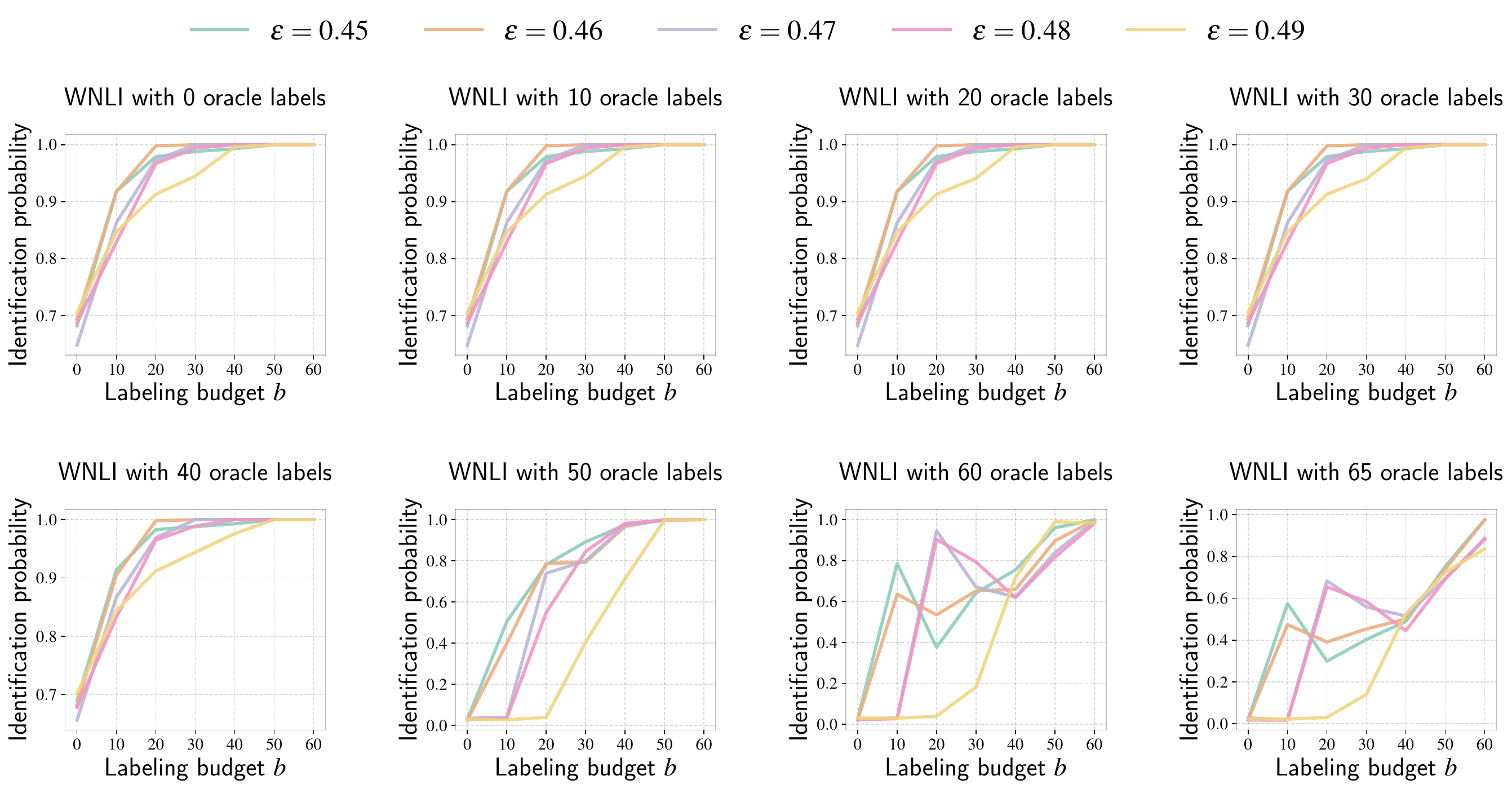} 
    \caption{Best model identification probability of \framework{} for $\epsilon \in [0.45, 49]$ on WNLI dataset, as the number of oracle labels increases.} 
    \label{fig:noisy_to_gt_wnli}
    \vspace{-0mm}
\end{figure*}

As discussed in \Cref{sec:epsilon}, it is a standard practice to allocate an initial budget for exploration~\citep{lewis1995sequential,mccallum1998employing,zhang2002active,hoi2006large,zhan2022comparative}.
In \Cref{fig:noisy_to_gt_rte} and \Cref{fig:noisy_to_gt_wnli} we show how identification probability changes as we increase the number of oracle labels for RTE and WNLI, respectively. For both RTE and WNLI the identification probability differs when evaluated on noisy labels and on oracle labels. 
As the number of oracle labels increases, the identification probability increasingly resembles that obtained when evaluated entirely on oracle labels.
However, the choice for best $\epsilon$ remains the same for both datasets when evaluated on noisy labels and on oracle labels, supporting the claim we do not any initial labeled data.

\clearpage

\subsection{Can We Use Noisy Labels to Identify the Best Model?} 
Although we can use the noisy oracle, constructed as explained in \Cref{sec:epsilon}, to choose the best $\epsilon$, we cannot use it to select the best model. \Cref{tab:best_model_acc_gap} shows the accuracy gap between the best model evaluated using the noisy oracle and the true best model. For all model collections, this accuracy gap is significantly greater than $0\%$. For example, the gap reaches up to $15.93\%$ for RTE.

\begin{table}[!htbp]
    \centering
\begin{tabular}{lc}
\specialrule{1.5pt}{0pt}{0pt}
Dataset & Best model accuracy gap \\
\midrule

CIFAR10-High & $4.28\%$ \\
CIFAR10-Low  &  $2.64\%$ \\
EmoContext   & $0.96\%$ \\
PACS         & $1.56\%$ \\
Drift        & $13.78\%$ \\
ImageNet     & $3.49\%$ \\
ImageNet Pytorch & $4.47\%$ \\
ImageNetV2 T-I & $4.53\%$ \\
ImageNetV2 T-0.7 & $5.72\%$ \\
ImageNetV2 M-F & $7.27\%$ \\
MRPC         & $1.29\%$ \\
CoLA         & $5.22\%$ \\
QNLI         & $3.25\%$ \\
QQP          & $1.08\%$ \\
SST-2        & $3.93\%$ \\
WNLI         & $3.49\%$ \\
MNLI         & $3.48\%$ \\
RTE          & $15.93\%$ \\
\specialrule{1.5pt}{0pt}{0pt}
\end{tabular}
    \caption{Accuracy gaps between the best models evaluated using the noisy oracle and the true best models across all model collections. Labels are required to identify the best model.}
    \label{tab:best_model_acc_gap}
\end{table}

\section{Extended Results}
\label{sec:app_extended_results}

In this section, we extend the results from \Cref{sec:experiments}.

\Cref{tab:label_eff_extended} presents the label efficiency for near-best models, as explained in \Cref{sec:near_best}, considering larger values of $\delta$. To achieve accuracy within the same $\delta$ vicinity of the best model, our method requires fewer labels than the best competing baselines (mostly \textsc{uncertainty} and \textsc{margin}).
For ImageNetV2 T-0.7, \framework{} reduces the labeling cost by $64.71\%$, $9.09\%$, and $16.67\%$ for $\delta$ values of $5\%$, $3\%$, and $2\%$, respectively. 
However, by examining the distribution of model accuracies in \Cref{fig:model_accuracies}, we observe that many models are within $5\%$, $3\%$, and $2\%$ of the best model's accuracy. Therefore, selecting a model within these vicinities cannot truly be considered selecting a near-best model, and we present these results for completeness. 
Importantly, as $\delta$ decreases, \framework{} achieves greater reductions in labeling costs.

\begin{table}[!htbp]
    \centering
    \vspace{-3mm}
\begin{tabular}{lccc}
\specialrule{1.5pt}{0pt}{0pt}
Dataset  & $\delta = 5\%$ & $\delta = 3\%$ & $\delta = 2\%$ \\
\midrule
CIFAR10-High & $\uparrow19.47\%$ & $\downarrow\textbf{20.99}\%$ & $\downarrow\textbf{25.81}\%$ \\
CIFAR10-Low & $\downarrow\textbf{8.18}\%$ & $\downarrow\textbf{15.79}\%$ & $\downarrow\textbf{19.29}\%$ \\
EmoContext & $\uparrow16.67\%$ & $\uparrow27.27\%$ & $\uparrow5.06\%$ \\
PACS & $\uparrow17.39\%$ & $\uparrow27.16\%$ & $\downarrow\textbf{49.80}\%$ \\
Drift & $\uparrow15.52\%$ & $\uparrow16.09\%$ & $\uparrow12.93\%$ \\
ImageNet & $\downarrow\textbf{35.00}\%$ & $\uparrow67.35\%$ & $\uparrow0.62\%$ \\
ImageNet Pytorch & $\uparrow33.33\%$ & $\uparrow10.94\%$ & $\uparrow11.94\%$ \\
ImageNetV2 T-I & $\uparrow18.75\%$ & $\downarrow\textbf{3.23}\%$ & $\downarrow\textbf{18.75}\%$ \\
ImageNetV2 T-0.7 & $\downarrow\textbf{64.71}\%$ & $\downarrow\textbf{9.09}\%$ & $\downarrow\textbf{16.67}\%$ \\
ImageNetV2 M-F & $\uparrow12.50\%$ & $\downarrow\textbf{28.09}\%$ & $\downarrow\textbf{13.48}\%$ \\
MRPC & $\uparrow2.90\%$ & $\uparrow5.33\%$ & $\uparrow2.60\%$ \\
CoLA & $\uparrow32.50\%$ & $\downarrow\textbf{9.60}\%$ & $\downarrow\textbf{41.99}\%$ \\
QNLI & $\downarrow\textbf{18.97}\%$ & $\downarrow\textbf{9.38}\%$ & $\downarrow\textbf{42.52}\%$ \\
QQP & $\uparrow41.73\%$ & $\uparrow62.58\%$ & $\uparrow35.22\%$ \\
SST-2 & $\uparrow59.26\%$ & $\uparrow14.81\%$ & $\downarrow\textbf{3.57}\%$ \\
WNLI & $\downarrow\textbf{1.59}\%$ & $\downarrow\textbf{1.59}\%$ & $0.00\%$ \\
MNLI & $\downarrow\textbf{29.27}\%$ & $\uparrow54.74\%$ & $\uparrow10.56\%$ \\
RTE & $\downarrow\textbf{40.96}\%$ & $\downarrow\textbf{40.96}\%$ & $\downarrow\textbf{40.96}\%$ \\
\specialrule{1.5pt}{0pt}{0pt}
\end{tabular}
    \caption{Label efficiency for near-best models: \framework{} consistently reduces labeling cost to reach the $\delta$ vicinity of the true best model compared to the best competing method as $\delta$ decreases.}
   \vspace{-0.2em}
    \label{tab:label_eff_extended}
\end{table}

\Cref{tab:percentile_combined_extended} extends the robustness analysis from \Cref{sec:robustness}. Instead of calculating the $95$th
percentile of the accuracy gap, we report the $90$th percentile of the accuracy gap across all realizations. As previously, we evaluate this for different budgets
for each dataset, determined as the budget required
for \framework{} to achieve $70\%$, $80\%$, $90\%$, and
$100\%$ identification probability. As shown in \Cref{tab:percentile_combined_extended}, \framework{} achieves
significantly smaller accuracy gaps compared to baseline methods. For example, 
the best competing
methods (\textsc{margin} and \textsc{uncertainty}) on the
RTE dataset with high disagreement among the
model predictions, for identification probability of
$70\%$, and $80\%$, return a model with $15.60\%$, and
$14.40\%$ accuracy gaps, while \framework{} returns models with accuracy gaps that are
$8.40\%$, and $6.80\%$. Quantitatively, these are
$1.9\times$ and $2.1\times$ smaller accuracy gaps.
Furthermore, when \framework{}s $90$th percentile accuracy gaps become $0\%$, the best competing baselines still select models
with significant accuracy gaps. For example,  \textsc{margin} and \textsc{uncertainty} select a model with $1.54\%$ accuracy gap for CoLA, and $6.40\%$ accuracy gap for RTE.

We confirm our findings highlight the robustness of \framework{} in consistently returning the near-best model, even at its lowest performance.

\begin{table*}[!htbp]
    \centering
    \resizebox{1\textwidth}{!}{%
\begin{tabular}{l c c c c c c}
\specialrule{1.5pt}{0pt}{0pt}
Dataset & \framework{} & \textsc{random} & \textsc{margin} & \textsc{uncertainty} & \textsc{amc} & \textsc{vma} \\
Identification probability & \small(70\%/80\%/90\%/100\%) & \small(70\%/80\%/90\%/100\%) & \small(70\%/80\%/90\%/100\%) & \small(70\%/80\%/90\%/100\%) & \small(70\%/80\%/90\%/100\%) & \small(70\%/80\%/90\%/100\%) \\
\midrule
CIFAR10-High & \textbf{1.00}/\textbf{0.40}/\textbf{0.10}/\textbf{0.00} & 4.20/2.90/2.60/2.10 & \underline{1.30}/\underline{0.70}/0.80/0.80 & 1.50/0.90/\underline{0.60}/\underline{0.40} & 3.10/1.80/1.60/1.30 & 3.30/2.10/1.70/1.30 \\
CIFAR10-Low & \textbf{1.00}/\textbf{0.50}/\textbf{0.10}/\textbf{0.00} & 1.40/1.20/1.00/0.90 & 1.60/1.40/1.30/1.00 & 1.60/1.40/1.20/0.90 & \underline{1.30}/\underline{1.00}/\underline{0.80}/\underline{0.70} & 1.40/1.10/1.00/0.80 \\
EmoContext & \textbf{0.60}/\textbf{0.30}/\textbf{0.10}/\textbf{0.00} & \underline{0.80}/0.70/0.60/0.40 & 1.00/\underline{0.60}/\underline{0.50}/\underline{0.30} & 0.90/0.70/\underline{0.50}/\underline{0.30} & 0.90/\underline{0.60}/\underline{0.50}/0.40 & \underline{0.80}/\underline{0.60}/\underline{0.50}/0.40 \\
PACS & \textbf{1.10}/\textbf{0.60}/\textbf{0.10}/\textbf{0.00} & 1.50/1.40/1.30/1.30 & 1.50/1.50/1.50/1.50 & \underline{1.30}/\underline{1.30}/\underline{1.20}/1.20 & 1.40/\underline{1.30}/\underline{1.20}/\underline{1.10} & 1.50/1.40/1.30/1.20 \\
Drift & \underline{8.67}/\textbf{6.40}/\underline{2.93}/\textbf{0.00} & \textbf{8.53}/\textbf{6.40}/3.87/\textbf{0.00} & 15.73/15.73/12.80/\underline{7.07} & 16.93/16.40/9.33/9.20 & 9.47/6.80/4.00/\textbf{0.00} & 9.60/5.87/\textbf{0.00}/\textbf{0.00} \\
ImageNet & \textbf{0.70}/\textbf{0.70}/\textbf{0.20}/\textbf{0.00} & 2.00/1.60/1.70/1.70 & \underline{0.80}/\underline{0.80}/\underline{0.80}/\underline{0.70} & 0.90/1.10/1.10/0.90 & 1.30/1.30/1.10/1.10 & 1.40/1.40/1.50/1.40 \\
ImageNet Pytorch & \textbf{0.50}/\textbf{0.30}/\textbf{0.20}/\textbf{0.00} & 2.80/3.00/2.50/2.10 & 1.00/\underline{0.60}/\underline{0.60}/\underline{0.50} & \underline{0.80}/0.80/0.70/\underline{0.50} & 1.40/1.00/0.90/0.80 & 2.60/2.00/1.30/0.90 \\
ImageNetV2 T-I & \textbf{1.00}/\textbf{0.30}/\textbf{0.00}/\textbf{0.00} & 3.50/3.20/2.20/1.80 & \textbf{1.00}/1.10/0.70/0.40 & 1.10/\underline{0.90}/\underline{0.50}/\underline{0.20} & 2.70/2.20/1.80/1.20 & 2.40/1.60/1.80/1.00 \\
ImageNetV2 T-0.7 & \textbf{0.70}/\textbf{0.30}/\textbf{0.00}/\textbf{0.00} & 3.80/2.90/2.90/1.90 & \underline{1.10}/\underline{0.90}/\underline{0.80}/\underline{0.60} & \underline{1.10}/1.00/0.90/0.70 & 2.20/1.50/1.20/0.80 & 2.00/1.40/1.50/1.20 \\
ImageNetV2 M-F & \textbf{0.60}/\textbf{0.30}/\textbf{0.10}/\textbf{0.00} & 3.30/1.50/1.50/1.10 & \underline{0.90}/0.70/\underline{0.50}/\underline{0.40} & \underline{0.90}/\underline{0.60}/0.60/\underline{0.40} & 2.50/0.90/0.90/0.80 & 2.70/1.50/1.10/0.90 \\
MRPC & \textbf{0.86}/\textbf{0.57}/\textbf{0.29}/\textbf{0.00} & 4.86/4.57/4.00/4.00 & \underline{1.14}/\underline{0.86}/\underline{0.86}/\underline{0.86} & \underline{1.14}/\underline{0.86}/\underline{0.86}/\underline{0.86} & 4.57/4.29/4.00/4.00 & 4.57/4.29/4.00/3.43 \\
CoLA & \textbf{0.62}/\textbf{0.38}/\textbf{0.12}/\textbf{0.00} & 2.87/2.75/2.75/2.62 & \underline{0.75}/\underline{0.75}/\underline{0.87}/\underline{1.12} & \underline{0.75}/\underline{0.75}/0.88/\underline{1.12} & 2.25/2.25/2.12/2.00 & 2.12/2.13/2.00/1.88 \\
QNLI & \textbf{0.70}/\textbf{0.40}/\textbf{0.10}/\textbf{0.00} & 4.00/3.70/3.50/3.30 & \underline{1.40}/\underline{0.90}/0.70/\underline{0.60} & 1.90/\underline{0.90}/\underline{0.60}/\underline{0.60} & 4.20/3.60/3.30/3.20 & 3.90/3.60/3.30/3.00 \\
QQP & \underline{0.32}/\textbf{0.14}/\textbf{0.02}/\textbf{0.00} & 1.36/1.24/1.16/1.10 & \textbf{0.28}/\underline{0.18}/\underline{0.16}/\underline{0.14} & \textbf{0.28}/\underline{0.18}/\underline{0.16}/\underline{0.14} & 0.88/0.78/0.60/0.54 & 0.90/0.76/0.66/0.54 \\
SST-2 & \textbf{0.27}/\textbf{0.13}/\textbf{0.00}/\textbf{0.00} & 5.87/5.73/5.20/5.20 & \underline{0.40}/\underline{0.40}/\underline{0.40}/\underline{0.40} & \underline{0.40}/\underline{0.40}/\underline{0.40}/\underline{0.40} & 4.53/4.53/4.13/4.13 & 4.40/4.27/4.13/4.00 \\
WNLI & \textbf{3.08}/\textbf{1.54}/\textbf{1.54}/\textbf{0.00} & 9.23/\underline{3.08}/\textbf{1.54}/\underline{1.54} & \underline{4.62}/\textbf{1.54}/\textbf{1.54}/\textbf{0.00} & 6.15/\textbf{1.54}/\textbf{1.54}/\textbf{0.00} & 6.15/\underline{3.08}/\textbf{1.54}/\underline{1.54} & 6.15/\underline{3.08}/\textbf{1.54}/\textbf{0.00} \\
MNLI & \textbf{0.70}/\textbf{0.40}/\textbf{0.10}/\textbf{0.00} & 3.30/3.00/2.70/1.90 & \underline{0.90}/\underline{0.90}/0.90/\underline{0.80} & 1.00/\underline{0.90}/\underline{0.80}/\underline{0.80} & 3.20/2.90/2.20/1.80 & 3.20/3.30/2.60/2.10 \\
RTE & \textbf{8.40}/\textbf{6.80}/\textbf{0.00}/\textbf{0.00} & 18.80/17.20/12.00/10.00 & 16.00/15.60/15.60/\underline{6.40} & \underline{15.60}/15.60/15.60/\underline{6.40} & 19.20/15.60/11.60/10.00 & 18.40/\underline{14.40}/\underline{11.20}/10.00 \\

\specialrule{1.5pt}{0pt}{0pt}
\end{tabular}
    }
    \caption{Robustness analysis: $90$-th Percentile Accuracy Gap ($\%$) at budget needed for \framework{} to reach identification probability $70\%, 80\%, 90\%$, and $100\%$. Compared to baselines, \framework{} achieves a smaller accuracy gap from the best model.
    Best method bolded; Next best underlined.}
    \label{tab:percentile_combined_extended}
    \vspace{-.9em}
\end{table*}

\section{Scaling and Computation Cost}
\label{sec:app_resources}
All methods, including \framework{} and the baseline algorithms, are implemented in Python and use the following Python libraries: Pandas, Matplotlib, Numpy, Scipy, Torch, HuggingFace, and
Seaborn. To optimize runtime, we execute the realizations in parallel on a $128$-core cluster.

The total runtime for the baseline methods across all realizations ranges from $60$ seconds (for WNLI) to $30$ hours (for ImageNet) on the cluster. \framework{} execution times range from $1.1$ seconds (for WNLI) to $61.2$ minutes (for ImageNet).
\end{document}